\documentclass{article} 
\usepackage{nips15submit_e,times}

\usepackage{xcolor}
\definecolor{mydarkblue}{rgb}{0,0.15,0.7}
\usepackage[colorlinks, urlcolor=mydarkblue, citecolor=mydarkblue, linkcolor=mydarkblue]{hyperref}

\usepackage{graphicx} 
\usepackage{subfigure} 
\usepackage{float}
\usepackage{verbatim}

\usepackage{amsmath}
\usepackage{amssymb}
\usepackage{amsfonts,amsthm}

\usepackage[numbers]{natbib}
\usepackage{natbibspacing}

\usepackage{algorithm}
\usepackage{algorithmic}

\newcommand{\myparagraph}[1]{\textbf{#1}\quad}

\DeclareMathOperator*{\argmax}{\arg\!\max}
\newcommand{\eqdef}{\overset{\mathrm{def}}{=\joinrel=}}

\newcommand{\avsum}{\mathop{\mathpalette\avsuminner\relax}\displaylimits}
\makeatletter
\newcommand\avsuminner[2]{%
  {\sbox0{$\m@th#1\sum$}%
   \vphantom{\usebox0}%
   \ooalign{%
     \hidewidth
     \smash{\vrule height\dimexpr\ht0+1pt\relax depth\dimexpr\dp0+1pt\relax}%
     \hidewidth\cr
     $\m@th#1\sum$\cr
   }%
  }%
}
\makeatother
\newcommand{\wsum}{\avsum}

\newcommand{\w}{\tau}

\newcommand{\pa}[0]{\mathrm{pa}}

\newcommand{\vv}[1]{\boldsymbol{#1}}

\usepackage{amsthm}   
\newtheorem{thm}{Theorem}[section]

\newtheorem{pro}[thm]{Proposition}    
\begingroup
    \makeatletter
    \@for\theoremstyle:=definition,remark,plain\do{%
        \expandafter\g@addto@macro\csname th@\theoremstyle\endcsname{%
            \addtolength\thm@preskip\parskip
            }%
        }
\endgroup

\title{Decomposition Bounds for Marginal MAP}

\author{
Wei Ping$^\ast$ \quad \quad \ \  Qiang Liu$^\dagger$  \quad \ \  Alexander Ihler$^\ast$ \\
$^\ast$Computer Science, UC Irvine\quad \ \ $^\dagger$Computer Science, Dartmouth College  \\
\texttt{ \{wping,ihler\}@ics.uci.edu \ qliu@cs.dartmouth.edu}
}

\nipsfinalcopy 

\begin{document}

\maketitle

\begin{abstract}
Marginal MAP inference involves making MAP predictions in systems defined with latent variables 
or missing information. It is significantly more difficult than pure marginalization and MAP tasks, 
for which a large class of efficient and convergent variational algorithms, such as dual decomposition, exist.
In this work, we generalize dual decomposition to a generic \emph{power sum} inference task, 
which includes marginal MAP, along with pure marginalization and MAP, as special cases. 
Our method is based on a block coordinate descent algorithm on a new convex decomposition bound, 
that is guaranteed to converge monotonically, and can be parallelized efficiently.  
We demonstrate our approach on marginal MAP queries defined 
on real-world problems from the UAI approximate
inference challenge, showing that our framework is faster and more reliable than previous methods.
\end{abstract}

\section{Introduction}
Probabilistic graphical models such as Bayesian networks and Markov random
fields provide a useful framework and powerful tools for machine learning.
Given a graphical model, \emph{inference} refers to answering probabilistic
queries about the model.  There are three common types of inference tasks.  The
first are max-inference or maximum a \emph{posteriori} (MAP) tasks, which aim
to find the most probable state of the joint probability; exact and approximate MAP 
inference is widely used in structured prediction \citep{nowozin11}.
Sum-inference tasks include calculating marginal
probabilities and the normalization constant of the distribution, and 
play a central role in many learning tasks (e.g., maximum likelihood).
Finally, marginal MAP tasks are ``mixed'' inference problems,
 which generalize the first two types by marginalizing a subset of variables (e.g., hidden variables) 
before optimizing over the remainder.%
\footnote{In some literature \citep[e.g.,][]{park2004complexity}, marginal MAP is simply called MAP, and the joint MAP task is called MPE.} 
These tasks arise in latent variable models \citep[e.g.,][]{ping14, Naradowsky12} and many decision-making
problems \citep[e.g.,][]{kiselev14}.
%
All three inference types are generally intractable; as a result, 
approximate inference, particularly convex relaxations or upper bounding
methods, are of great interest. 

Decomposition methods provide a useful and computationally efficient class of bounds
on inference problems.  For example, dual decomposition methods for MAP
\citep[e.g.,][]{sontag11} give a class of easy-to-evaluate upper bounds which
can be directly optimized using coordinate descent \citep{werner07,globerson08}, subgradient updates \citep{komodakis11},
or other methods \citep[e.g.,][]{meshi11}.  It is easy to ensure both convergence, and that the objective 
is monotonically decreasing (so that more computation always provides a better bound).
The resulting bounds can be used either as stand-alone approximation methods \citep{globerson08,komodakis11},
or as a component of search \citep{ihler12a}.  
In summation problems, a notable decomposition bound is tree-reweighted BP (TRW), 
which bounds the partition function with a combination of trees 
\citep[e.g.,][]{wainwright05, meltzer2009convergent, jancsary11, domke11}.
These bounds are useful in joint inference and learning
(or ``inferning'') frameworks, allowing learning with approximate inference to be framed
as a joint optimization over the model parameters and decomposition bound, often leading
to more efficient learning \citep[e.g.,][]{meshi10}.
However, far fewer methods have been developed for marginal MAP problems.

In this work, we deveop a decomposition bound that has a number of desirable properties:
(1) \emph{Generality}: our bound is sufficiently general to be applied easily to marginal MAP.
(2) \emph{Any-time}: it yields a bound at any point during the optimization (not just at convergence), 
so it can be used in an any-time way.
(3) \emph{Monotonic and convergent}: more computational effort gives strictly tighter bounds; 
note that (2) and (3) are particularly
important for high-width approximations, which are expensive to represent and update.
(4) Allows \emph{optimization over all parameters}, including the ``weights'', or fractional counting numbers,
of the approximation; these parameters often have a significant effect on the tightness of the resulting bound.
(5) \emph{Compact representation}: within a given class of bounds, using fewer parameters to express the bound
reduces memory and typically speeds up optimization.

We organize the rest of the paper as follows. 
Section~\ref{sec:notation} gives some background and notation, 
followed by connections to related work in 
Section~\ref{sec:relatedwork}. 
We derive our decomposed bound in
Section~\ref{sec:DD_bound}, and present a (block) coordinate descent algorithm
for monotonically tightening it in 
Section~\ref{sec:Algorithm}.
We report experimental results in
Section~\ref{sec:experiments} and conclude the paper in
Section~\ref{sec:conclusion}.

\section{Background}
\label{sec:notation}
Here, we review some background on graphical models and inference tasks.
A Markov random field (MRF) on discrete random variables $x= [x_1,\ldots, x_n] \in \mathcal{X} ^n$ is a probability distribution, 
\begin{align}
\label{factor-graph-model}
p(x; \theta) = \exp \Big[ \sum_{\alpha \in \mathcal{F}} \theta_{\alpha} (x_\alpha) - \Phi(\theta) \Big];
\quad
\Phi(\theta) = \log \sum_{x\in \mathcal{X}^n} \exp \Big[ \sum_{\alpha \in \mathcal{F}} \theta_{\alpha} (x_\alpha)  \Big],
\end{align}
where $\mathcal{F}$ is a set of subsets of the variables, each associated with a factor $\theta_\alpha$,
and $\Phi(\theta)$ is the log partition function.
We associate an undirected graph $G = (V, E)$ with $p(x)$ by mapping each $x_i$ to a node $i\in V$, and adding 
an edge $ij\in E$ iff there exists $\alpha \in \mathcal{F}$ such that $\{i, j\} \subseteq \alpha$. 
We say node $i$ and $j$ are neighbors if $ij \in E$. Then, $\mathcal{F}$ is the subset of cliques 
(fully connected subgraphs) of $G$.

The use and evaluation of a given MRF often involves different types of inference tasks. 
\emph{Marginalization}, or \emph{sum-inference} tasks perform a sum over the configurations 
to calculate the log partition function $\Phi$ in \eqref{factor-graph-model}, marginal probabilities, 
or the probability of some observed evidence.
On the other hand, the maximum \emph{a posteriori} (MAP), or \emph{max-inference} tasks perform joint 
maximization to find configurations with the highest probability, that is,
$\Phi_{0}(\theta) = \max_{x}  \sum_{\alpha \in \mathcal{F}} \theta_{\alpha} (x_\alpha) $.

A generalization of max- and sum- inference is \emph{marginal MAP}, 
or \emph{mixed-inference},
in which we are interested in first marginalizing a subset $A$ of variables (e.g., hidden variables),
and then maximizing the remaining variables $B$ (whose values are of direct interest), that is,
\begin{align}
\label{marginal_map}
\Phi_{AB}(\theta) = \max_{x_B} Q(x_B) = \max_{x_B}  \log \sum_{x_A} \exp \Big[ \sum_{\alpha \in \mathcal{F}} \theta_{\alpha} (x_\alpha)  \Big],
\end{align}
where $A\cup B = V$ (all the variables) and $A\cap B = \emptyset$. Obviously,
both sum- and max- inference are special cases of marginal MAP when $A =  V$ and $B = V$, respectively.

It will be useful to define an even more general inference task, based on a power sum operator:
\begin{align*}
\wsum_{x_i}^{\w_i} f(x_i) = \big[ \sum_{x_i}  {f(x_i)}^ {1/\w_i}  \big ] ^{\w_i},
%
\end{align*}
where $f(x_i)$ is any non-negative function and $\w_i$ is a \emph{temperature} or \emph{weight} parameter.
The power sum reduces to a standard sum when $\w_i = 1$, and approaches $\max_x f(x)$ when $\w_i \to 0^+$,
so that we define the power sum with $\w_i = 0$ to equal the max operator.

The power sum is helpful for unifying max- and sum- inference~\citep[e.g.,][]{Weiss07MAP}, as well as marginal MAP~\cite{liu14}. 
Specifically, we can apply power sums with different weights $\w_i$ to each variable $x_i$ along a predefined 
elimination order (e.g., $[x_1, \ldots, x_n]$), to define the \emph{weighted log partition function}:
\begin{align}
\label{powered-sum-inference}
\Phi_{\vv \w} (\theta) 
 = \log  \wsum_x^{\vv \w} \exp (\theta(x))  
 = \log \wsum_{x_n}^{\w_n}  \dots\wsum_{x_1}^{\w_1} \exp (\theta(x)), 
\end{align}
where we note that the value of \eqref{powered-sum-inference} depends on the elimination order unless all the weights are equal.
Obviously, \eqref{powered-sum-inference} includes marginal MAP \eqref{marginal_map} as a special case by 
setting weights ${\vv \w}_A = 1$ and ${\vv \w}_B=0$.
This representation provides a useful tool for understanding and deriving new 
algorithms for general inference tasks, especially marginal MAP, for which relatively few efficient algorithms exist.

\section{Related Work}
\label{sec:relatedwork}
Variational upper bounds on MAP and the partition function, along with algorithms for providing fast, convergent 
optimization, have been widely studied in the last decade.
In MAP, dual decomposition and linear programming methods have become a dominating approach, 
with numerous optimization techniques \cite{werner07,globerson08,sontag09, komodakis11, yarkony2010covering, ruozzi2013, meshi11}, 
and methods to tighten the approximations \cite{sontag08,komodakis11}. 

For summation problems, most upper bounds are derived from the tree-reweighted (TRW) family of convex bounds \cite{wainwright05},
or more generally conditional entropy decompositions \cite{globerson07}.
TRW bounds can be framed as optimizing over a convex combination of tree-structured models, or in a dual
representation as a message-passing, TRW belief propagation algorithm.
This illustrates a basic tension in the resulting bounds: in its primal form
\footnote{Despite the term ``dual decomposition'' used in MAP tasks, in this work we refer to decomposition bounds as ``primal''
bounds, since they can be viewed as directly bounding the result of variable elimination.  This is in contrast
to, for example, the linear programming relaxation of MAP, which bounds the result only after optimization.}
(combination of trees),
TRW is inefficient: it maintains a weight and $O(|V|)$ parameters for each tree, 
and a large number of trees may be required to obtain a tight bound;
this uses memory and makes optimization slow.  
On the other hand, the dual, or free energy, form uses only $O(|E|)$ parameters (the TRW messages) to optimize over the 
set of all possible spanning trees -- but, the resulting optimization is only guaranteed to be a bound at
convergence, 
\footnote{See an example in Supplement \ref{sec:ising}.} 
making it difficult to use in an anytime fashion.  Similarly, the gradient of the weights 
is only correct at convergence, making it difficult to optimize over these parameters; most
implementations \cite[e.g.,][]{libdai} simply adopt fixed weights.

Thus, most algorithms do not satisfy all the desirable properties listed in the introduction. 
For example, many works have developed convergent message-passing algorithms for convex free energies
\cite[e.g.,][]{hazan08, hazan10}.  However, by optimizing the dual they do not provide
a bound until convergence, and the representation and constraints on the counting numbers do not facilitate
optimizing the bound over these parameters.  
To optimize counting numbers, \cite{hazan_peng2012} adopt a more restrictive free energy form 
requiring positive counting numbers on the entropies; but this cannot represent marginal MAP,  
whose free energy involves conditional entropies (equivalent to the difference between two entropy terms).

On the other hand, working in the primal domain ensures a bound, but usually at the cost of 
enumerating a large number of trees.  \cite{jancsary11} heuristically select a small number of
trees to avoid being too inefficient, while \cite{meltzer2009convergent} focus on trying to speed up
the updates on a given collection of trees.
Another primal bound is weighted mini-bucket (WMB, \cite{liu11}), which can represent a large 
collection of trees compactly and is easily applied to marginal MAP 
using the weighted log partition function viewpoint \cite{liu14,marinescu14}; 
however, existing optimization algorithms for WMB are non-monotonic, and often fail to converge,
especially on marginal MAP tasks.

While our focus is on variational bounds \cite{liu11,liu13}, 
there are many non-variational approaches for marginal MAP as well.
\citep{park03uai, yuan09ijcai}~provide upper bounds on marginal MAP by reordering the order in which
variables are eliminated, and using exact inference in the reordered join-tree; however, this is
exponential in the size of the (unconstrained) treewidth, and can easily become intractable.
\citep{meek2011}~give an approximation closely related to mini-bucket \citep{dechter03} to bound 
the marginal MAP; however, unlike (weighted) mini-bucket, these bounds cannot be improved iteratively.
The same is true for the algorithm of \citep{maua2012anytime}, which also has a strong dependence on treewidth.
Other examples of marginal MAP algorithms 
include local search \citep[e.g.,][]{park2004complexity} and 
Markov chain Monte Carlo methods \citep[e.g.,][]{doucet02,yuan04}.

\vspace{-0.1em}
\section{Fully Decomposed Upper Bound}
\label{sec:DD_bound}
\vspace{-0.1em}
In this section, we develop a new general form of upper bound 
and provide an efficient, monotonically convergent optimization algorithm.
Our new bound is based on fully decomposing the graph into disconnected cliques,  
allowing very efficient local computation, but can still be as tight as 
WMB or the TRW bound with a large collection of spanning trees 
once the weights and shifting variables are chosen or optimized properly. 
Our bound reduces to dual decomposition for MAP inference, but is applicable to more general mixed-inference settings. 

Our main result is based on the following generalization of the classical H{\"o}lder's inequality~\citep{hardy52}: 
\vspace{-0.5em}
\begin{thm}
\label{thm:main}
For a given graphical model $p(x ; \theta)$ in \eqref{factor-graph-model} with cliques $\mathcal{F} = \{\alpha\}$ and a set of 
non-negative weights ${\vv \w} = \{ \w_i \geq 0, i\in V \}$, 
we define a set of ``{split weights}" ${\bf w^\alpha} = \{ w_i^{\alpha} \geq 0,  i \in \alpha \}$ on each variable-clique pair $(i, \alpha)$, that satisfies 
$\sum_{ \alpha | \alpha \ni i } w_i^{\alpha} = \w_i. $
Then we have
\vspace{-0.7em}
\begin{align}
\label{holder_ieq}
& \wsum_x^{\vv \w} \prod_{\alpha \in \mathcal{F} }  \exp \big[ \theta_\alpha (x_\alpha) \big] \le 
\prod_{\alpha \in \mathcal{F} }  \wsum_{x_\alpha}^{\bf w^\alpha} \exp \big[ \theta_\alpha (x_\alpha) \big], 
\end{align}
where the left-hand side is the powered-sum along order $[x_1,\ldots, x_n]$ as defined in \eqref{powered-sum-inference}, and the right-hand side is the product of the powered-sums on subvector $x_{\alpha}$ with weights ${\bf w}^\alpha$ along the same elimination order; that is,
$
\wsum_{x_\alpha}^{\bf w^\alpha} \exp\big[\theta_\alpha (x_\alpha)\big]  
=  \wsum_{x_{k_c}}^{w_{k_c}^{\alpha}} \cdots  \wsum_{x_{k_1}}^{w_{k_1}^{\alpha}}
\exp \big[ \theta_\alpha  (x_\alpha) \big] ,
$
where $x_{\alpha} = [x_{k_1}, \ldots, x_{k_c}]$ should be ranked with increasing index, consisting with the elimination order $[x_1,\ldots, x_n]$ as used in the left-hand side. 
\end{thm}
\vspace{-0.5em}
Proof details can be found in Section \ref{sec:proof_4.1} 
of the supplement. A key advantage 
of the bound \eqref{holder_ieq} 
is that it decomposes the joint power sum on $x$ into a product of independent power sums over smaller 
cliques $x_{\alpha}$, which significantly reduces computational complexity and enables parallel computation.

\vspace{-0.1em}
\subsection{Including Cost-shifting Variables}
\vspace{-0.1em}
In order to increase the flexibility of the upper bound, we introduce a set of 
\emph{cost-shifting} or \emph{reparameterization} 
variables ${\delta} = \{ \delta_i^{\alpha}(x_i) ~|~ \forall (i,\alpha), i\in \alpha  \}$ on each variable-factor pair $(i, \alpha)$, which can be optimized to provide a much tighter upper bound. 
Note that $\Phi_{\vv \w} (\theta)$ can be rewritten as,
\vspace{-0.5em}
\begin{align*}
{\small
\Phi_{\vv \w}(\theta) =  \log \wsum_x^{\vv \w} \exp \Big[ \sum_{i \in V} \sum_{\alpha \in N_i}  \delta_i^{\alpha}(x_i) 
 + \sum_{\alpha \in \mathcal{F}} \big( \theta_{\alpha} (x_\alpha) - \sum_{i \in \alpha} \delta_i^{\alpha}(x_i) \big) \Big],
 }
\end{align*}
where $N_i = \{ \alpha ~|~ \alpha \ni i \}$ is the set of cliques incident to $i$. 
Applying inequality \eqref{holder_ieq}, we have that
\vspace{-0.4em}
\begin{align}
\label{bound_wsum}
{\small 
\!\!\!\!
\Phi_{\vv \w}(\theta) 
\le \sum_{i \in V} \log\wsum_{x_i}^{w_i} \exp \Big[  \sum_{\alpha \in N_i} \delta_i^{\alpha}(x_i) \Big] 
+ \sum_{\alpha \in \mathcal{F} } \log\wsum_{x_\alpha}^{\bf w^\alpha } 
			\exp \Big[\theta_{\alpha} (x_\alpha) -\sum_{i\in \alpha} \delta_i^{\alpha}(x_i) \Big] 
\eqdef   L(\delta,  {\bf w}) 
}, 
\end{align}
where the nodes $i\in V$ are also treated as cliques within inequality \eqref{holder_ieq}, 
and a new weight $w_i$ is introduced on each variable $i$;
the new weights 
${\bf w} = \{ w_i, w_i^\alpha ~|~ \forall(i,\alpha), i\in\alpha \}$ should satisfy
\vspace{-0.4em}
\begin{align}
\label{weight_bound_wsum}
{\small
\ w_i + \sum_{\alpha \in N_i} w_i^{\alpha} = \w_i,\ \ w_i \ge 0, \   w_i^{\alpha} \ge 0,\ \ \forall (i, \alpha). 
}
\end{align}
The bound $L(\delta,  {\bf w})$ 
 is convex w.r.t.\ the cost-shifting variables $\delta$ and weights ${\bf w}$, 
enabling an efficient optimization algorithm that we present in Section~\ref{sec:Algorithm}. 
As we will discuss in Section~\ref{sec:gradient}, these shifting variables correspond to Lagrange multipliers that enforce a moment matching condition.

\vspace{-0.1em}
\subsection{Dual Form and Connection With Existing Bounds}
\label{sec:dual}
\vspace{-0.1em}
It is straightforward to see that our bound in \eqref{bound_wsum} reduces to dual decomposition \citep{sontag11} when applied on MAP inference with all $\w_i = 0$, and hence $w_i = w^{\alpha}_i = 0$. 
On the other hand, its connection with sum-inference bounds such as WMB and TRW 
is seen more clearly via a dual representation of \eqref{bound_wsum}:
\begin{thm}
\label{thm:dual}
The tightest upper bound obtainable by \eqref{bound_wsum}, that is,
\vspace{-0.05em}
\begin{align}
\label{dual_form}
\min_{\bf w} \min_{\delta} L(\delta, {\bf w}) 
= \min_{\bf w} \max_{ {\bf b} \in \mathbb{L}(G)} 
 \big \{  
 			\langle \theta, b \rangle + \sum_{i \in V} w_i H(x_i ; b_i) 
 			+  \sum_{\alpha \in \mathcal{F}} \sum_{i \in \alpha}  w^{\alpha}_i H(x_i | x_{{\pa}^{\alpha}_i} ~;~b_{\alpha}) 
 \big\}, 
\end{align}
where ${\bf b} = \{b_i(x_i) , b_{\alpha}(x_{\alpha}) ~|~ \forall (i,\alpha), i\in \alpha \}$ is a set of pseudo-marginals (or beliefs) defined on the singleton variables and the cliques, and $\mathbb{L}$ is the corresponding local consistency polytope defined by 
$\mathbb{L}(G) = \{ {\bf b} ~|~b_i(x_i)= \sum_{x_{\alpha \setminus i}} b_{\alpha}(x_{\alpha}), ~ \sum_{x_i} b_i(x_i)=1 \}$.
Here, $H(\cdot )$ are their corresponding marginal or conditional entropies, 
and $\pa^{\alpha}_i$ is the set of variables 
in $\alpha$ that rank later than $i$, that is, for the global elimination order $[x_1, \ldots, x_n]$, $\pa^{\alpha}_i = \{  j \in \alpha ~ | ~  j \succ i \}$.
\end{thm}
\vspace{-0.5em}
The proof details can be found in Section~\ref{sec:dual_repres} 
 of the supplement.
It is useful to compare Theorem~\ref{thm:dual} with other dual representations.
As the sum of non-negatively weighted conditional entropies, 
the bound is clearly convex and within the general class of conditional entropy decompositions~(CED)~\cite{globerson07}, 
but unlike generic CED it has a simple and efficient primal form \eqref{bound_wsum}.
\footnote{The primal form derived in \cite{globerson07} (a geometric program) is computationally infeasible. }
Comparing to the dual form of WMB
in Theorem~4.2 of \cite{liu11}, our bound is as tight as WMB, and 
hence the class of TRW / CED bounds attainable by WMB \cite{liu11}.
%
Most duality-based forms \cite[e.g.,][]{hazan08,hazan10} are expressed in terms of joint entropies, 
$\langle \theta, b \rangle + \sum_\beta c_\beta H(b_\beta)$, rather than conditional entropies;
while the two can be converted, the resulting counting numbers
$c_\beta$ will be differences of weights $\{w_i^\alpha\}$,
\footnote{See more details of this connection in Section~\ref{subsec:connection_free_energy} 
 of the supplement.}
which obfuscates its convexity, 
makes it harder to maintain the relative constraints on the counting numbers during optimization, 
and makes some counting numbers negative (rendering some methods inapplicable \cite{hazan_peng2012}).
Finally, like most variational bounds in dual form, the RHS of \eqref{dual_form} has a inner maximization and hence guaranteed
to bound $\Phi_{\vv \w}(\theta)$ only at its optimum.

\begin{figure*}[tbp]
   \centering
   \begin{tabular}{cccc}
   \includegraphics[height=.145\textwidth]{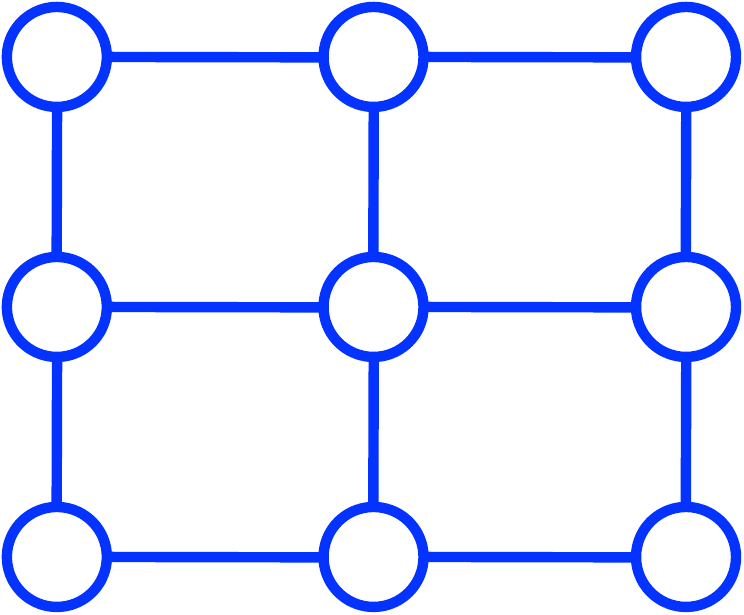} &
   \includegraphics[height=.145\textwidth]{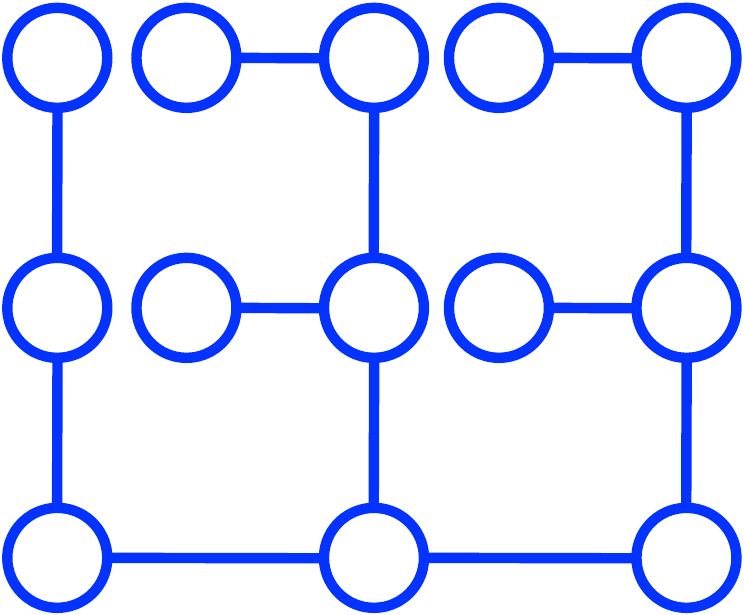} &
     \includegraphics[height=.145\textwidth]{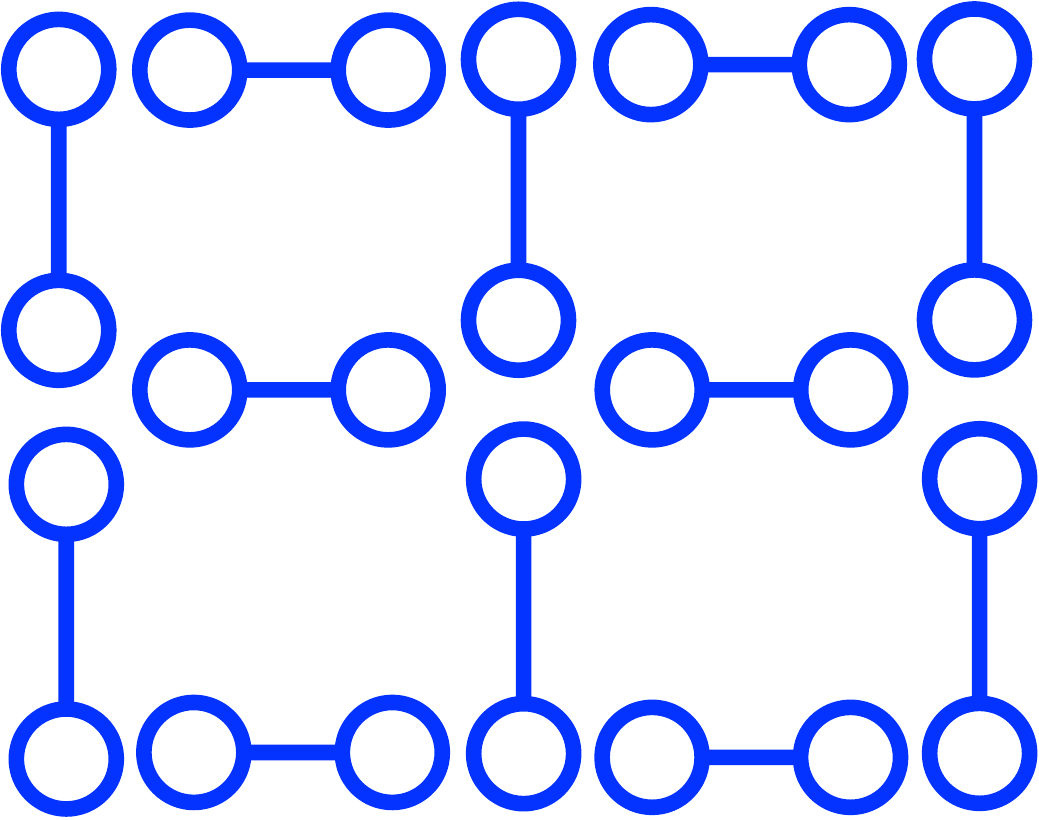} &
   \includegraphics[height=.145\textwidth]{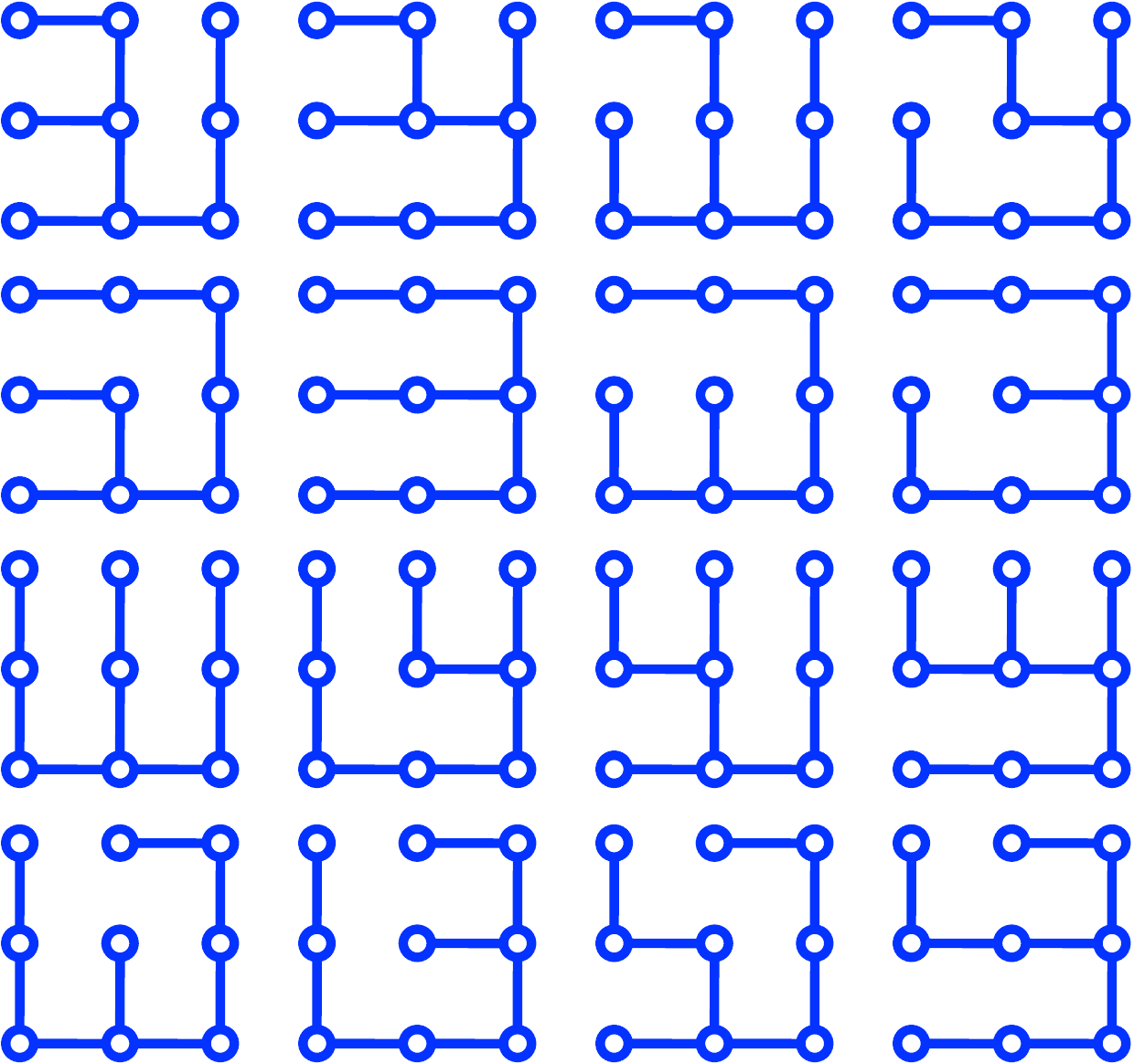}\\ 
   (a) $3\times 3$ grid & (b) WMB: covering tree  & (c) Full decomposition  & (d) TRW
   \end{tabular}
   \caption{Illustrating WMB, TRW and our bound on (a) $3\times3$ grid. (b) WMB uses a covering tree 
with a minimal number of splits and cost-shifting. (c) Our decomposition 
\eqref{bound_wsum} further splits the graph into small cliques (here, edges), introducing 
additional cost-shifting variables but allowing for easier, monotonic optimization.
(d) Primal TRW splits the graph into many
spanning trees, requiring even more cost-shifting variables. 
Note that all three bounds attain the same tightness after optimization.}
   \label{fig:gridexample}
\vspace{-0.5em}
\end{figure*}

In contrast, our Eq.~\eqref{bound_wsum} is a primal bound (hence, a bound for any $\delta$).  It is
similar to the primal form of TRW, except that (1) the individual regions are single cliques, rather than
spanning trees of the graph, 
\footnote{While non-spanning subgraphs can be used in the primal TRW form, doing so leads to 
loose bounds; in contrast, our decomposition's terms consist of individual cliques.}
%
and (2) the fraction weights ${\bf w^\alpha}$ associated with each region are
vectors, rather than a single scalar.
The representation's efficiency can be seen with an example in Figure~\ref{fig:gridexample},
which shows a $3\times 3$ grid model and three relaxations that achieve the same bound.
Assuming $d$ states per variable and ignoring the equality constraints, 
our decomposition in Figure~\ref{fig:gridexample}(c) 
uses $24d$ cost-shifting parameters ($\delta$), and $24$ weights.
WMB (Figure~\ref{fig:gridexample}(b)) 
is slightly more efficient, with only $8d$ parameters for $\delta$ and and $8$ weights, but
its lack of decomposition makes parallel and monotonic updates difficult.  On the other hand,
the equivalent primal TRW uses 16 spanning trees, 
shown in Figure~\ref{fig:gridexample}(d),
for $16\cdot 8\cdot d^2$ parameters, and 16 weights.
The increased dimensionality of the optimization slows convergence, and updates are non-local,
requiring full message-passing sweeps on the involved trees 
(although this cost can be amortized in some cases \cite{meltzer2009convergent}).

\vspace{-0.1em}
\section{Monotonically Tightening the Bound}
\label{sec:Algorithm}
\vspace{-0.1em}
In this section, we propose a block coordinate descent algorithm (Algorithm~\ref{alg:GDD}) 
to minimize the upper bound  $L(\delta, {\bf w})$ in \eqref{bound_wsum} 
w.r.t. the shifting variables $\delta$ and weights ${\bf w}$. 
Our algorithm has a monotonic convergence property, 
and allows efficient, distributable local computation due to the full decomposition of our bound. 
Our framework allows generic powered-sum inference, including
max-, sum-, or mixed-inference as special cases by setting different weights. 

\vspace{-0.1em}
\subsection{Moment Matching and Entropy Matching}
\label{sec:gradient}
\vspace{-0.1em}
We start with deriving the gradient of $L(\delta, {\bf w})$ w.r.t.\ $\delta$ and ${\bf w}$. 
We show that the zero-gradient equation w.r.t. $\delta$ has a simple form of moment matching that enforces a consistency between the \emph{singleton beliefs} with their related \emph{clique beliefs}, and that of weights ${\bf w}$ enforces a consistency of \emph{marginal} and \emph{conditional entropies}.
\vspace{-0.4em}
\begin{thm}
\label{thm:matching}
(1) For $L(\delta, {\bf w})$ in \eqref{bound_wsum}, its zero-gradient w.r.t.\ $\delta_i^\alpha(x_i)$ is 
\vspace{-0.4em}
\begin{align}
\label{moment_matching}
\frac{ \partial L } { \partial \delta_i^{\alpha} (x_i) } =  \mu_i (x_i)  - \sum_{ x_{\alpha \backslash i} } \mu_{\alpha}(x_\alpha)
= 0,
\end{align}
where $ \mu_i (x_i)  \propto \exp \big[  \frac{1}{w_i}{ \sum_{\alpha \in N_i} \delta_i^{\alpha}(x_i)  }  \big] $ can be interpreted as a singleton belief on $x_i$, 
and $\mu_{\alpha} (x_{\alpha})$ can be viewed as clique belief on $x_{\alpha}$,
defined with a chain rule (assuming $x_{\alpha} = [x_1, \ldots, x_{c}]$),  
$
\mu_{\alpha} (x_{\alpha}) = \prod_{i=1}^{c} \mu_{\alpha} (x_i | x_{i+1:c}); 
\
\mu_{\alpha} (x_i | x_{i+1:c}) =  (  { Z_{i-1} (x_{i:c})  } / { Z_i (x_{i+1:c})  } )^{ 1 / w_i^\alpha },
$
where $Z_i$ is the partial powered-sum up to $x_{1:i}$ on the clique, that is, 
\vspace{-0.4em}
\begin{align*}
Z_i (x_{i+1:c}) = \wsum_{x_i}^{w_i^\alpha} \cdots \wsum_{x_1}^{w_1^\alpha} 
\exp \Big[\theta_{\alpha} (x_\alpha) -\sum_{i\in \alpha} \delta_i^{\alpha}(x_i) \Big], \quad
Z_0 (x_\alpha)  = \exp \Big[\theta_{\alpha} (x_\alpha) -\sum_{i\in \alpha} \delta_i^{\alpha}(x_i) \Big], 
\end{align*}
where the summation order should be consistent with the global elimination order  ${\bf o}=[x_1,\ldots,  x_n]$. 

(2) The gradients of $L(\delta, {\bf w})$  w.r.t. the weights $\{w_i, w_i^\alpha \}$ are marginal and conditional entropies defined on  the beliefs $\{\mu_i, \mu_{\alpha}\}$, respectively, 
\vspace{-0.2em}
\begin{align}
\label{gradient_weight}
\!\!\!\!\!\!\!\!\!\!\!\!\!\!\!\!
\frac{ \partial L } { \partial w_i } = H(x_i; \mu_i), \quad\quad 
\frac{ \partial L } { \partial w_i^{\alpha} } = H(x_i| x_{i+1:c} ; \mu_\alpha)
= - \sum_{x_\alpha}  \mu_\alpha(x_\alpha) \log \mu_\alpha(x_i | x_{i+1:c}).
\end{align}
Therefore, the optimal weights should satisfy the following KKT condition
\vspace{-0.2em}
\begin{align}
\label{entropy_matching}
w_i \big( H(x_i; \mu_i) - \bar{H_i} \big) = 0,  \quad
w_i^\alpha \big( H(x_i| x_{i+1:c} ; \mu_\alpha) - \bar{H_i} \big) = 0, \quad
\forall (i, \alpha)
\end{align}
where $\bar{H_i}  = \big( w_i H(x_i; \mu_i) + \sum_{\alpha} w_i^\alpha H(x_i| x_{i+1:c} ; \mu_\alpha) \big)/\w_i$ is the average entropy on node $i$.
\end{thm}
The proof details can be found in Section~\ref{sec:Proof_Theorem5.1} 
 of the supplement.
The matching condition \eqref{moment_matching} enforces that
$\mu = \{\mu_i, \mu_{\alpha} ~|~ \forall (i,\alpha)\}$ belong to the local consistency 
polytope $\mathbb{L}$ as defined in Theorem~\ref{thm:dual}; similar moment matching results 
appear commonly in variational inference algorithms \citep[e.g.,][]{wainwright05}. 
\cite{wainwright05} also derive a gradient of the weights, but it is based on the free energy form 
and is correct only after optimization; 
our form holds at any point, enabling efficient joint optimization of $\delta$ and ${\bf w}$.

\begin{algorithm}[tb]
   \caption{Generalized Dual-decomposition (GDD)}
   \label{alg:GDD}
\begin{algorithmic}
 \STATE {\bfseries Input:} 
   		weights $\{\w_i ~|~ i\in V\}$, elimination order ${\bf o}$.
 \STATE {\bfseries Output:}
     the optimal $\delta^*, {\bf w^*}$ giving tightest upper bound L($\delta^*, {\bf w^*}$) 
     for $\Phi_{\w}(\theta)$ in \eqref{bound_wsum}.
     
 \vspace{0.3em}
 \STATE initialize $\delta = 0$ and weights ${\bf w} = \{ w_i, w_i^\alpha \}$.
\REPEAT
   \FOR{node $i$ (in parallel with node j, $(i,j) \not\in E$)  }
   		\IF{$\w_i = 0$}
   			\STATE update ${\vv \delta}_{N_i} = \{\delta_i^{\alpha} |   \forall \alpha \in N_i\}$ with the closed-form update \eqref{closed-star-update}; 
    	\ELSIF{$\w_i \neq 0$ }
     		\STATE update  ${\vv \delta}_{N_i}$ and ${\bf w}_{N_i}$ with gradient descent \eqref{moment_matching} and\eqref{weight_update_rule}, combined with line search;	
    	\ENDIF   
   \ENDFOR
\UNTIL{convergence}
 \STATE $\delta^* \leftarrow \delta$,  ${\bf w^*} \leftarrow {\bf w}$, 
 		and evaluate L($\delta^*, {\bf w^*}$) by \eqref{bound_wsum}; \\
 \vspace{.3em}
 \STATE \emph{Remark}. GDD solves max-, sum- and mixed-inference by setting different values of weights $\{\w_i\}$. 
\end{algorithmic}
\vspace{-0.3em}
\end{algorithm}

\vspace{-0.1em}
\subsection{Block Coordinate Descent}
\label{sec:algorithm}
\vspace{-0.1em}
We derive a block coordinate descent method in Algorithm~\ref{alg:GDD} to minimize our bound, in which we sweep through all the nodes $i$ and update each block ${\vv \delta}_{N_i} = \{ \delta_i^{\alpha}(x_i)\ | \ \forall  \alpha \in N_i\}$ 
and ${ {\bf w}_{N_i} = \{ w_i, w_i^\alpha ~|~\forall \alpha \in N_i \} }$
with the neighborhood parameters  fixed. 
Our algorithm applies two update types, depending on whether the variables have zero weight: 
(1) For nodes with $\w_i = 0$ (corresponding to max nodes $i \in B$ in marginal MAP), we derive a closed-form coordinate descent rule for the associated shifting variables ${\vv \delta}_{N_i}$;
these nodes do not require to optimize ${\bf w}_{N_i}$ since it is fixed to be zero.
(2) For nodes with $\w_i \neq 0$ (e.g., sum nodes $i \in A$ in marginal MAP), 
we lack a closed form update for ${\vv \delta}_{N_i}$ and ${\bf w}_{N_i} $, 
and optimize by local gradient descent combined with line search. 

The lack of a closed form coordinate update for nodes $\w_i \neq 0$ is mainly
because the order of power sums with different weights cannot be exchanged. 
However, the gradient descent inner loop is still efficient,
because each gradient evaluation only involves the local variables in clique $\alpha$.

\myparagraph{Closed-form Update.}
For any node $i$ with $\w_i = 0$ (i.e., max nodes $i\in B$ in marginal MAP), and its associated $ {\vv \delta}_{N_i} = \{ \delta_i^{\alpha}(x_i)\ | \ \forall  \alpha \in N_i\}$, 
the following update gives a closed form solution for the zero (sub-)gradient equation in \eqref{moment_matching} 
(keeping the other $\{ \delta_j^{\alpha} | j\neq i , \forall \alpha \in N_i \}$ fixed): 
\vspace{-1em}
\begin{align}
\label{closed-star-update}
\delta_i^{\alpha}(x_i) \leftarrow &\ 
\frac{|N_i|}{|N_i| + 1} \gamma_i^{\alpha}(x_i)  
- \frac{1}{|N_i| +1 }  \!\!\!  \sum_{\beta\in N_i \backslash \alpha} \!\!\! \gamma_i^{\beta}(x_i), 
\quad\qquad
\end{align}
\vspace{-0.25em}$\!\!\!$
where $|N_i|$ is the number of neighborhood cliques, and 
$\gamma_i^{\alpha}(x_i) = \log \wsum_{x_{\alpha \backslash i} }^{ {\bf w}_{\backslash i}^{\alpha}}  
\exp \big[ \theta_{\alpha}(x_\alpha) - \sum_{j \in \alpha \backslash i } \delta_j^{\alpha}(x_j) \big].$
Note that the update in \eqref{closed-star-update} 
works regardless of the weights of nodes $\{\w_j  ~|~ \forall  j \in \alpha,\ \forall \alpha \in N_i \}$ in the neighborhood cliques; when all the neighboring nodes also have zero weight ($\w_j = 0$ for $\forall j\in \alpha,\  \forall \alpha \in N_i$), it is analogous to 
the ``star'' update of dual decomposition for MAP~\citep{sontag11}.
The detailed derivation is shown in Proposition~\ref{pro_close-form-update} and~\ref{pro_close-form-star-update}
in the supplement. 

The update in \eqref{closed-star-update} can be calculated with a cost of only $O(|N_i|\cdot d^{|\alpha|})$, 
where $d$ is the number of states of $x_i$, and $|\alpha|$ is the clique size, by computing and 
saving all the shared $\{ \gamma_i^{\alpha}(x_i) \}$ before updating $\vv{\delta}_{N_i}$. 
Furthermore,
the updates of $\vv{\delta}_{N_i}$ for different nodes $i$ are independent if they are not directly 
connected by some clique $\alpha$; this makes it easy to parallelize the coordinate descent process by 
partitioning the graph into independent sets, and parallelizing the updates within each set.

\myparagraph{Local Gradient Descent.}
For nodes with $\w_i \neq 0$ (or $i\in A$ in marginal MAP),  
there is no closed-form solution for $\{ \delta_i^{\alpha}(x_i) \}$ and $\{ w_i,  w_i^\alpha \}$ to minimize the upper bound. 
However, because of the fully decomposed form, 
the gradient w.r.t.\ $\vv{\delta}_{N_i}$ and ${\bf w}_{N_i}$,
\eqref{moment_matching}--\eqref{gradient_weight}, can be evaluated efficiently 
via local computation with $O(|N_i|\cdot d^{|\alpha|})$, 
and again can be parallelized between nonadjacent nodes.
To handle the normalization constraint~\eqref{weight_bound_wsum} on ${\bf w}_{N_i}$,
we use an exponential gradient descent: 
let $w_i = \exp(v_i) / \big[ \exp(v_i) + \sum_{\alpha} \exp(v_i^\alpha) \big]$ 
and $w_i^\alpha = \exp(v_i^\alpha) / \big[ \exp(v_i) + \sum_{\alpha} \exp(v_i^\alpha) \big]$; 
taking the gradient w.r.t.\ $v_i$ and $v_i^\alpha$ and transforming back gives the following update
\vspace{-0.25em}
\begin{align}
\label{weight_update_rule}
\!\!\!
w_i &\propto w_i \exp \big[ -\eta w_i \big( H(x_i; \mu_i) - \bar{H_i}   \big) \big],
\ \ \ 
w_i^\alpha \propto  w_i^\alpha \exp \big[ -\eta w_i^\alpha \big(  H(x_i | x_{\pa_i^\alpha}; \mu_\alpha) - \bar{H_i}  \big)  \big], 
\end{align}
where $\eta$ is the step size and $\pa^{\alpha}_i \!=\! \{  j\! \in\! \alpha \, | \,  j \!\succ\! i \}$.
In our implementation, we find that 
a few gradient steps (e.g., 5) 
with a backtracking line search using the Armijo rule works well in practice. 
Other more advanced 
optimization methods, such as L-BFGS and Newton's method 
are also applicable.

\vspace{-0.1em}
\section{Experiments}
\label{sec:experiments}
\vspace{-0.4em}
In this section, we demonstrate our algorithm on a set of real-world graphical models from  recent UAI inference challenges, 
including two diagnostic Bayesian networks
with 203 and 359 variables and max 
domain sizes $7$ and $6$, respectively,
and several MRFs for pedigree analysis with up to $1289$ variables, max 
domain size of $7$ and clique size $5$.%
\footnote{See \url{http://graphmod.ics.uci.edu/uai08/Evaluation/Report/Benchmarks}.}
We construct marginal MAP problems on these models by randomly selecting 
half
of the variables to be max nodes, 
and the rest as sum nodes.

%

We implement several algorithms that optimize the same primal marginal MAP bound, 
including our GDD (Algorithm~\ref{alg:GDD}), the WMB algorithm in \citep{liu11} with $ibound=1$, 
which uses the same cliques and a fixed point heuristic for optimization, 
and an off-the-shelf L-BFGS implementation that directly optimizes our decomposed bound. 
For comparison, we also computed several related primal bounds, including 
standard mini-bucket \cite{dechter03}
and elimination reordering \cite{park03uai, yuan09ijcai}, limited to the same
computational limits ($ibound=1$).
We also tried MAS \cite{meek2011} but found its bounds extremely loose.%
\footnote{The instances tested have many zero probabilities, which make finding lower bounds difficult; since MAS' bounds 
are symmetrized, this likely contributes to its upper bounds being loose.}

Decoding (finding a configuration $\hat x_B$) is more difficult in marginal MAP than in joint MAP.
We use the same local decoding procedure that is standard in dual decomposition \cite{sontag11}.
However, evaluating the objective $Q(\hat x_B)$ involves a potentially difficult sum over $x_A$, making it hard to score
each decoding.  For this reason, we evaluate the score of each decoding, but show the most recent decoding
rather than the best (as is standard in MAP) to simulate behavior in practice.

\begin{figure*}[tb] 
\centering
\begin{tabular}{cc}
\!\!\!\!\!\!
\includegraphics[width=4.6cm, clip]{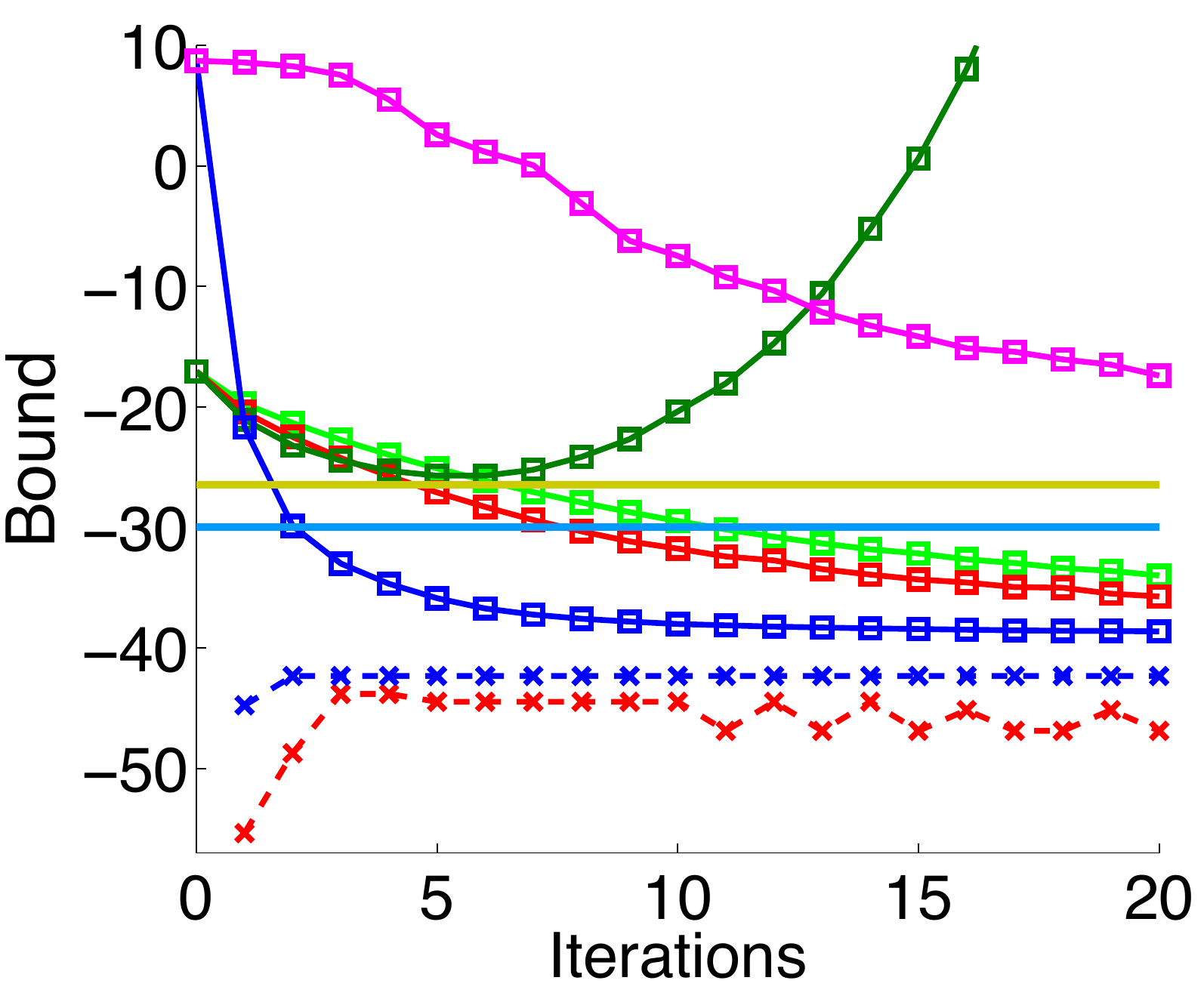} 
\hspace{-.7em}\raisebox{5em}{\includegraphics[width=2.35cm, clip]{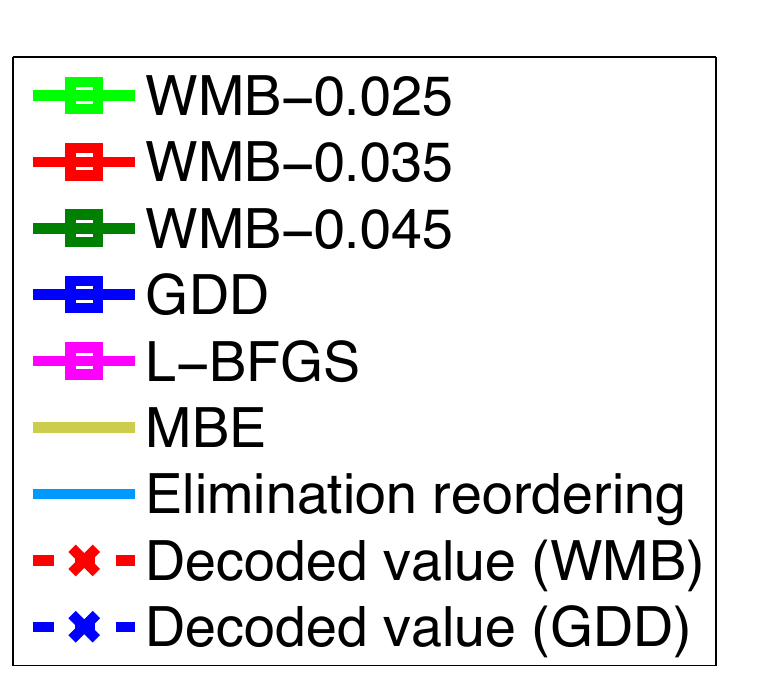}}
&
\includegraphics[width=4.6cm, clip]{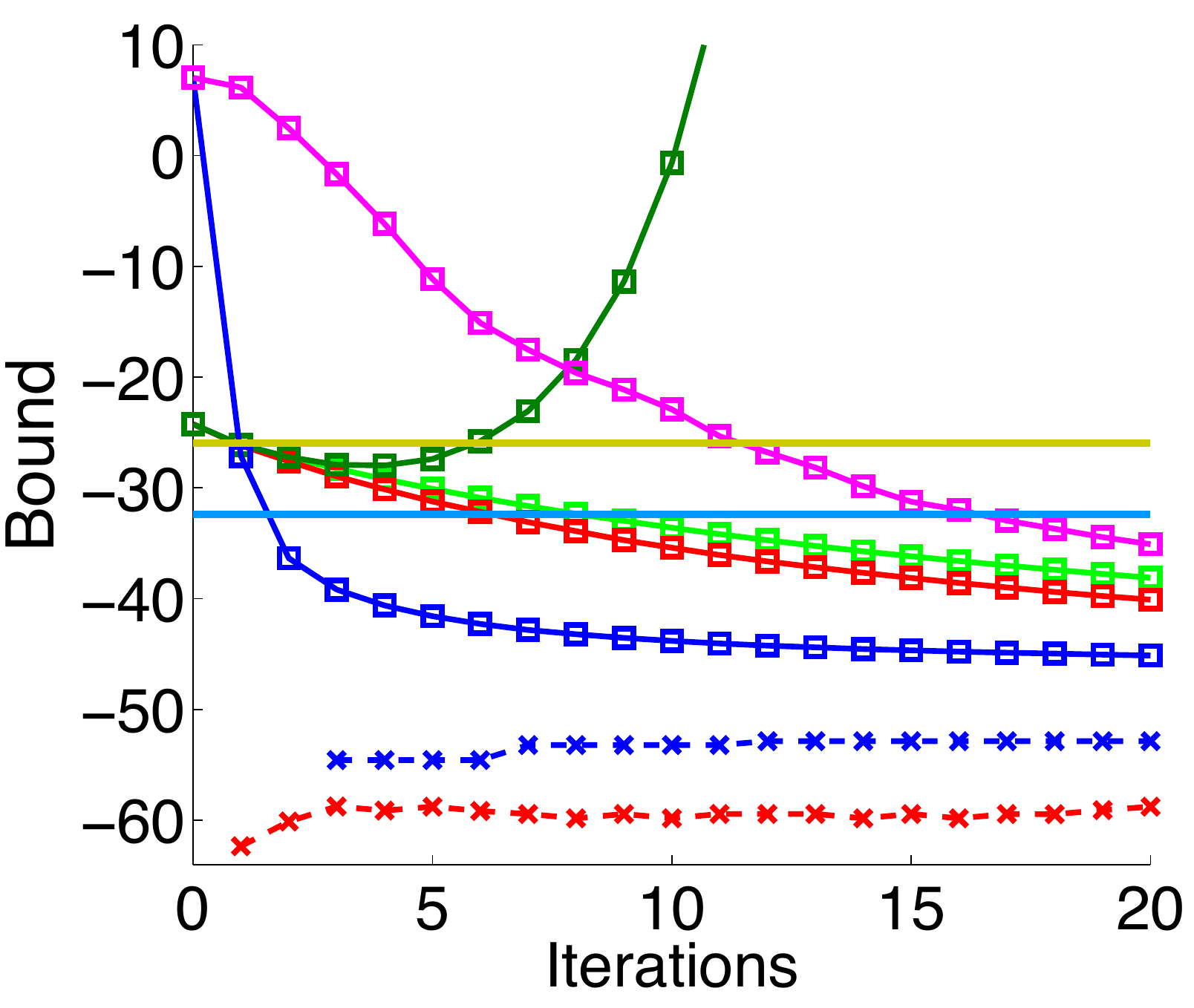}  
\hspace{-.7em}\raisebox{5em}{\includegraphics[width=2.35cm, clip]{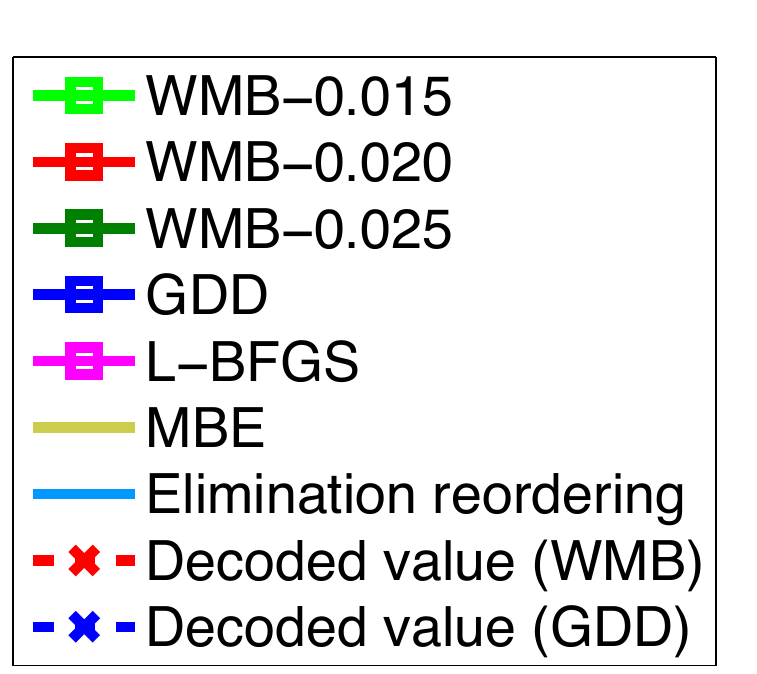} }
\\
{ \small (a) BN-1 (203 nodes) \qquad \qquad \qquad } 
& {\small (b) BN-2 (359 nodes) \qquad \qquad \qquad}  \\
\end{tabular}
\vspace{-0.9em}
\caption{ Marginal MAP results on BN-1 and BN-2 with $50\%$ randomly selected max-nodes~(additional plots are in the supplement~\ref{sec:diagnostic_BN}).  
We plot the upper bounds of different algorithms across iterations; 
the objective function $Q(x_B)$~\eqref{marginal_map} of the decoded solutions $x_B$ are also shown (dashed lines). 
At the beginning, $Q(x_B)$ may equal to $-\infty$ because of zero probabiliy.  
}
\vspace{0.2em}
\label{fig:DN}
\end{figure*}

\begin{figure*}[tb] \centering
\begin{tabular}{ccc}
\!\!\!\!\!
\includegraphics[width=4.4cm, clip]{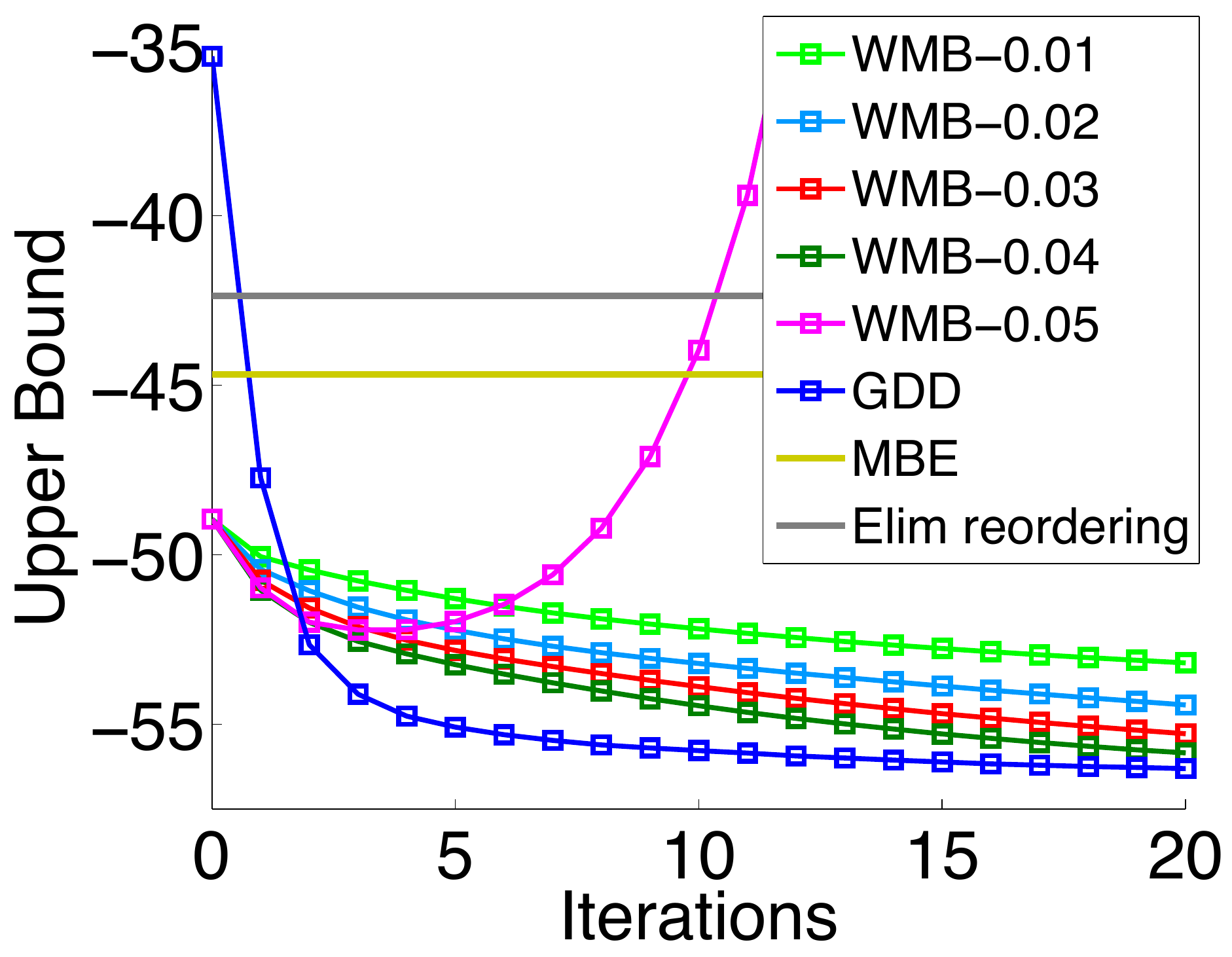} & \!\!\!
\includegraphics[width=4.4cm, clip]{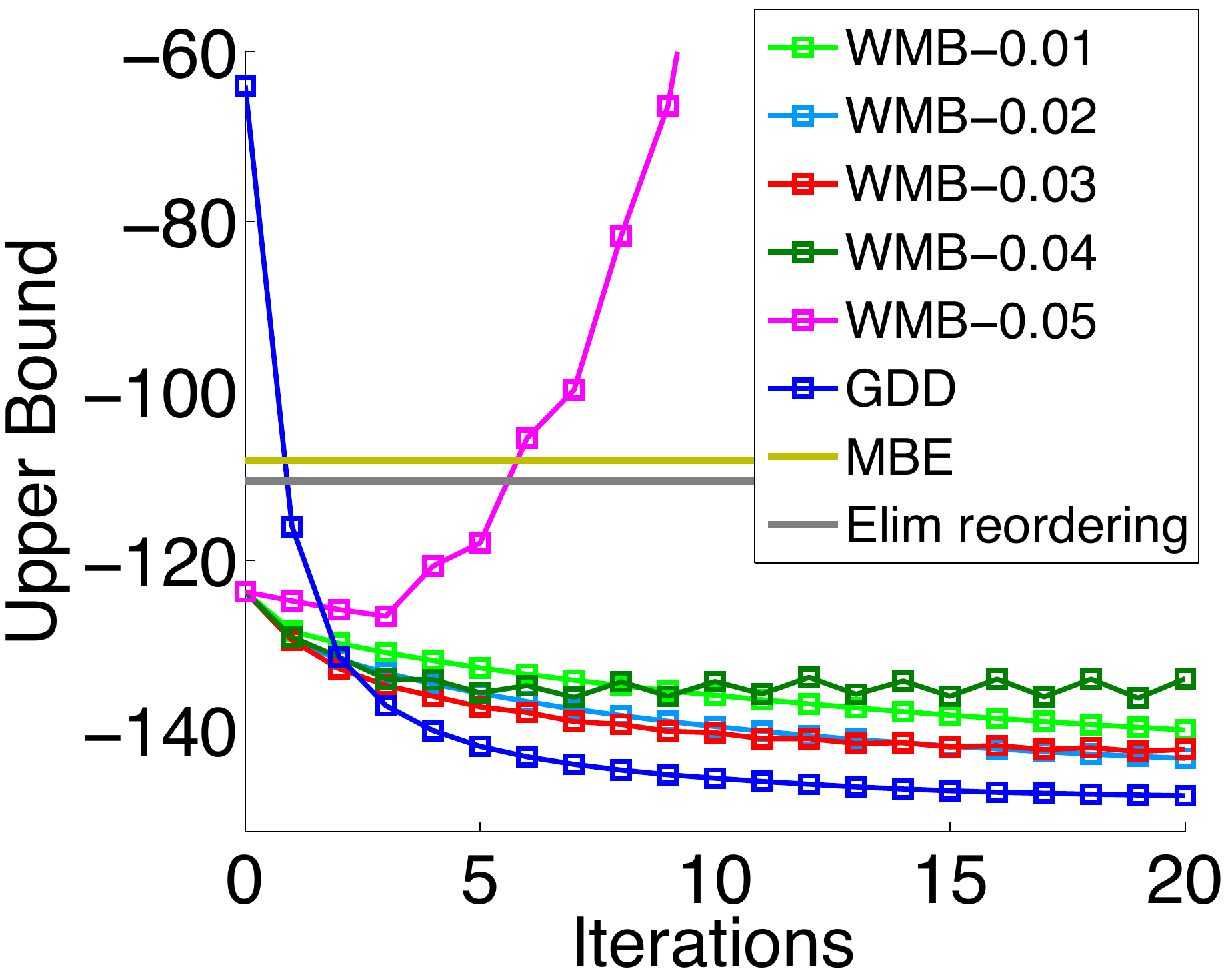} & \!\!\!
\includegraphics[width=4.4cm, clip]{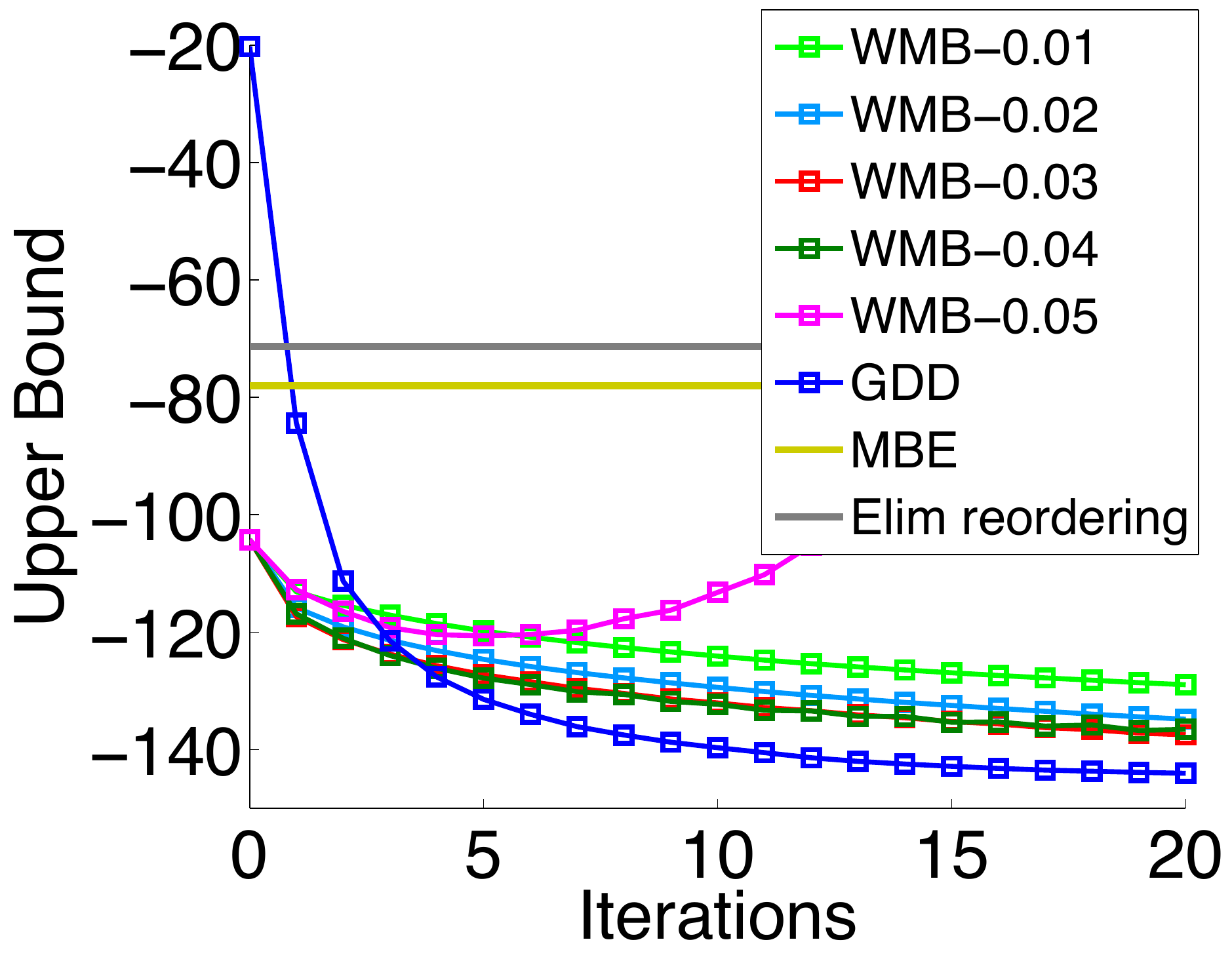}  
\\
{ \small (a) pedigree1 (334 nodes)} & {\small (b) pedigree7 (1068 nodes)}  & {\small (c) pedigree9 (1118 nodes)} 
\end{tabular}
\vspace{-0.9em}
\caption{ Marginal MAP inference on three pedigree models (additional plots are in the supplement~\ref{sec:pedigree}). 
We randomly select half the nodes as max-nodes in these models. 
We tune the damping rate of WMB from 0.01 to 0.05.
}
\label{fig:Link_margMAP} %
\vspace{-0.3em}
\end{figure*}

Figure~\ref{fig:DN} and Figure~\ref{fig:Link_margMAP} compare the convergence of the different algorithms, 
where we define the iteration of each algorithm to correspond to a full sweep over the graph, 
with the same order of time complexity: 
one iteration for GDD is defined in Algorithm~\ref{alg:GDD}; 
for WMB is a full forward and backward message pass, as in Algorithm~2 of \citep{liu11}; 
and for L-BFGS is a joint quasi-Newton step on all variables.
The elimination order that we use is obtained by a weighted-min-fill heuristic 
\citep{dechter2013reasoning} constrained to eliminate the sum nodes first.

\myparagraph{Diagnostic Bayesian Networks.}
Figure \ref{fig:DN}(a)-(b) shows that our GDD converges quickly and monotonically on both the networks, while WMB does not converge without proper damping;
we experimented different damping ratios for WMB, and found that it is slower than GDD even with the best damping ratio found (e.g., in Figure~\ref{fig:DN}(a), WMB works best with damping ratio $0.035$ (WMB-0.035), but is still significantly slower than GDD). 
Our GDD also gives better decoded marginal MAP solution $x_B$ (obtained by rounding the singleton beliefs). 
Both WMB and our GDD provide a much tighter bound than the non-iterative mini-bucket elimination (MBE) \citep{dechter03}
or reordered elimination \citep{park03uai, yuan09ijcai} methods.

\vspace{-0.1em}
\myparagraph{Genetic Pedigree Instances.}
Figure~\ref{fig:Link_margMAP} shows similar results on a set of pedigree instances.
Again, GDD outperforms WMB even with the best possible damping, 
and out-performs the non-iterative bounds after only one iteration (pass through the graph).

\vspace{-0.1em}
\section{Conclusion}
\label{sec:conclusion}
\vspace{-0.3em}
In this work, we propose a new class of decomposition bounds for general
powered-sum inference, which is capable of representing a large class of primal
variational bounds but is much more computationally efficient.
Unlike previous primal sum bounds, our bound decomposes into computations on small, local cliques, 
increasing efficiency and enabling parallel and monotonic optimization.
We derive a block
coordinate descent algorithm for optimizing our bound over both the cost-shifting parameters (reparameterization)
and weights (fractional counting numbers), 
which generalizes dual decomposition and enjoy similar monotonic convergence property. 
Taking the advantage of its monotonic convergence, our new algorithm can be widely applied as a 
building block for improved heuristic construction in search, or more efficient learning algorithms.  

\vspace{-0.1em}
\subsubsection*{Acknowledgments}
\vspace{-0.4em}
This work is sponsored in part by
NSF grants IIS-1065618 and IIS-1254071.
Alexander Ihler is also funded in part 
by the United States  Air Force under Contract No. FA8750-14-C-0011 
under the DARPA PPAML program.

\newpage 
\renewcommand{\bibsection}{\subsubsection*{References}} \small
\bibliographystyle{abbrv}
\bibliography{nips_decomp} 

\begin{thebibliography}{10}

\bibitem{dechter2013reasoning}
R.~Dechter.
\newblock Reasoning with probabilistic and deterministic graphical models:
  Exact algorithms.
\newblock {\em Synthesis Lectures on Artificial Intelligence and Machine
  Learning}, 2013.

\bibitem{dechter03}
R.~Dechter and I.~Rish.
\newblock Mini-buckets: A general scheme for bounded inference.
\newblock {\em JACM}, 2003.

\bibitem{domke11}
J.~Domke.
\newblock Dual decomposition for marginal inference.
\newblock In {\em AAAI}, 2011.

\bibitem{doucet02}
A.~Doucet, S.~Godsill, and C.~Robert.
\newblock Marginal maximum a posteriori estimation using {M}arkov chain {M}onte
  {C}arlo.
\newblock {\em Statistics and Computing}, 2002.

\bibitem{globerson07}
A.~Globerson and T.~Jaakkola.
\newblock Approximate inference using conditional entropy decompositions.
\newblock In {\em AISTATS}, 2007.

\bibitem{globerson08}
A.~Globerson and T.~Jaakkola.
\newblock Fixing max-product: Convergent message passing algorithms for {MAP}
  {LP}-relaxations.
\newblock In {\em NIPS}, 2008.

\bibitem{hardy52}
G.~H. Hardy, J.~E. Littlewood, and G.~Polya.
\newblock {\em Inequalities}.
\newblock Cambridge University Press, 1952.

\bibitem{hazan_peng2012}
T.~Hazan, J.~Peng, and A.~Shashua.
\newblock Tightening fractional covering upper bounds on the partition function
  for high-order region graphs.
\newblock In {\em UAI}, 2012.

\bibitem{hazan08}
T.~Hazan and A.~Shashua.
\newblock Convergent message-passing algorithms for inference over general
  graphs with convex free energies.
\newblock In {\em UAI}, 2008.

\bibitem{hazan10}
T.~Hazan and A.~Shashua.
\newblock Norm-product belief propagation: Primal-dual message-passing for
  approximate inference.
\newblock {\em IEEE Transactions on Information Theory}, 2010.

\bibitem{ihler12a}
A.~Ihler, N.~Flerova, R.~Dechter, and L.~Otten.
\newblock Join-graph based cost-shifting schemes.
\newblock In {\em UAI}, 2012.

\bibitem{jancsary11}
J.~Jancsary and G.~Matz.
\newblock Convergent decomposition solvers for {TRW} free energies.
\newblock In {\em AISTATS}, 2011.

\bibitem{kiselev14}
I.~Kiselev and P.~Poupart.
\newblock Policy optimization by marginal {MAP} probabilistic inference in
  generative models.
\newblock In {\em AAMAS}, 2014.

\bibitem{komodakis11}
N.~Komodakis, N.~Paragios, and G.~Tziritas.
\newblock {MRF} energy minimization and beyond via dual decomposition.
\newblock {\em TPAMI}, 2011.

\bibitem{liu14}
Q.~Liu.
\newblock {\em Reasoning and Decisions in Probabilistic Graphical Models--A
  Unified Framework}.
\newblock PhD thesis, University of California, Irvine, 2014.

\bibitem{liu11}
Q.~Liu and A.~Ihler.
\newblock Bounding the partition function using {H}$\ddot{o}$lder's inequality.
\newblock In {\em ICML}, 2011.

\bibitem{liu13}
Q.~Liu and A.~Ihler.
\newblock Variational algorithms for marginal {MAP}.
\newblock {\em JMLR}, 2013.

\bibitem{marinescu14}
R.~Marinescu, R.~Dechter, and A.~Ihler.
\newblock {AND/OR} search for marginal {MAP}.
\newblock In {\em UAI}, 2014.

\bibitem{maua2012anytime}
D.~Maua and C.~de~Campos.
\newblock Anytime marginal maximum a posteriori inference.
\newblock In {\em ICML}, 2012.

\bibitem{meek2011}
C.~Meek and Y.~Wexler.
\newblock Approximating max-sum-product problems using multiplicative error
  bounds.
\newblock {\em Bayesian Statistics}, 2011.

\bibitem{meltzer2009convergent}
T.~Meltzer, A.~Globerson, and Y.~Weiss.
\newblock Convergent message passing algorithms: a unifying view.
\newblock In {\em UAI}, 2009.

\bibitem{meshi11}
O.~Meshi and A.~Globerson.
\newblock An alternating direction method for dual {MAP} {LP} relaxation.
\newblock In {\em ECML/PKDD}, 2011.

\bibitem{meshi10}
O.~Meshi, D.~Sontag, T.~Jaakkola, and A.~Globerson.
\newblock Learning efficiently with approximate inference via dual losses.
\newblock In {\em ICML}, 2010.

\bibitem{libdai}
J.~Mooij.
\newblock lib{DAI}: A free and open source {C++} library for discrete
  approximate inference in graphical models.
\newblock {\em JMLR}, 2010.

\bibitem{Naradowsky12}
J.~Naradowsky, S.~Riedel, and D.~Smith.
\newblock Improving {NLP} through marginalization of hidden syntactic
  structure.
\newblock In {\em EMNLP}, 2012.

\bibitem{nowozin11}
S.~Nowozin and C.~Lampert.
\newblock Structured learning and prediction in computer vision.
\newblock {\em Foundations and Trends in Computer Graphics and Vision}, 6,
  2011.

\bibitem{park03uai}
J.~Park and A.~Darwiche.
\newblock Solving {MAP} exactly using systematic search.
\newblock In {\em UAI}, 2003.

\bibitem{park2004complexity}
J.~Park and A.~Darwiche.
\newblock Complexity results and approximation strategies for {MAP}
  explanations.
\newblock {\em JAIR}, 2004.

\bibitem{ping14}
W.~Ping, Q.~Liu, and A.~Ihler.
\newblock Marginal structured {SVM} with hidden variables.
\newblock In {\em ICML}, 2014.

\bibitem{ruozzi2013}
N.~Ruozzi and S.~Tatikonda.
\newblock Message-passing algorithms: Reparameterizations and splittings.
\newblock {\em IEEE Transactions on Information Theory}, 2013.

\bibitem{sontag11}
D.~Sontag, A.~Globerson, and T.~Jaakkola.
\newblock Introduction to dual decomposition for inference.
\newblock {\em Optimization for Machine Learning}, 2011.

\bibitem{sontag09}
D.~Sontag and T.~Jaakkola.
\newblock Tree block coordinate descent for {MAP} in graphical models.
\newblock {\em AISTATS}, 2009.

\bibitem{sontag08}
D.~Sontag, T.~Meltzer, A.~Globerson, T.~Jaakkola, and Y.~Weiss.
\newblock Tightening {LP} relaxations for {MAP} using message passing.
\newblock In {\em UAI}, 2008.

\bibitem{wainwright05}
M.~Wainwright, T.~Jaakkola, and A.~Willsky.
\newblock A new class of upper bounds on the log partition function.
\newblock {\em IEEE Transactions on Information Theory}, 2005.

\bibitem{wainwright2008graphical}
M.~Wainwright and M.~Jordan.
\newblock Graphical models, exponential families, and variational inference.
\newblock {\em Foundations and Trends in Machine Learning}, 2008.

\bibitem{Weiss07MAP}
Y.~Weiss, C.~Yanover, and T.~Meltzer.
\newblock {MAP} estimation, linear programming and belief propagation with
  convex free energies.
\newblock In {\em UAI}, 2007.

\bibitem{werner07}
T.~Werner.
\newblock A linear programming approach to max-sum problem: A review.
\newblock {\em TPAMI}, 2007.

\bibitem{yarkony2010covering}
J.~Yarkony, C.~Fowlkes, and A.~Ihler.
\newblock Covering trees and lower-bounds on quadratic assignment.
\newblock In {\em CVPR}, 2010.

\bibitem{yedidia2005constructing}
J.~S. Yedidia, W.~T. Freeman, and Y.~Weiss.
\newblock Constructing free-energy approximations and generalized belief
  propagation algorithms.
\newblock {\em IEEE Transactions on Information Theory}, 2005.

\bibitem{yuan09ijcai}
C.~Yuan and E.~Hansen.
\newblock Efficient computation of jointree bounds for systematic map search.
\newblock {\em IJCAI}, 2009.

\bibitem{yuan04}
C.~Yuan, T.~Lu, and M.~Druzdzel.
\newblock Annealed {MAP}.
\newblock In {\em UAI}, 2004.

\end{thebibliography}


\maketitle

\appendix

\section*{\Large Supplement}

\section{Experiment on Ising grid}
\label{sec:ising}
\begin{figure*}[th*] \centering
\begin{tabular}{cc}
\includegraphics[width=4.5cm, clip]{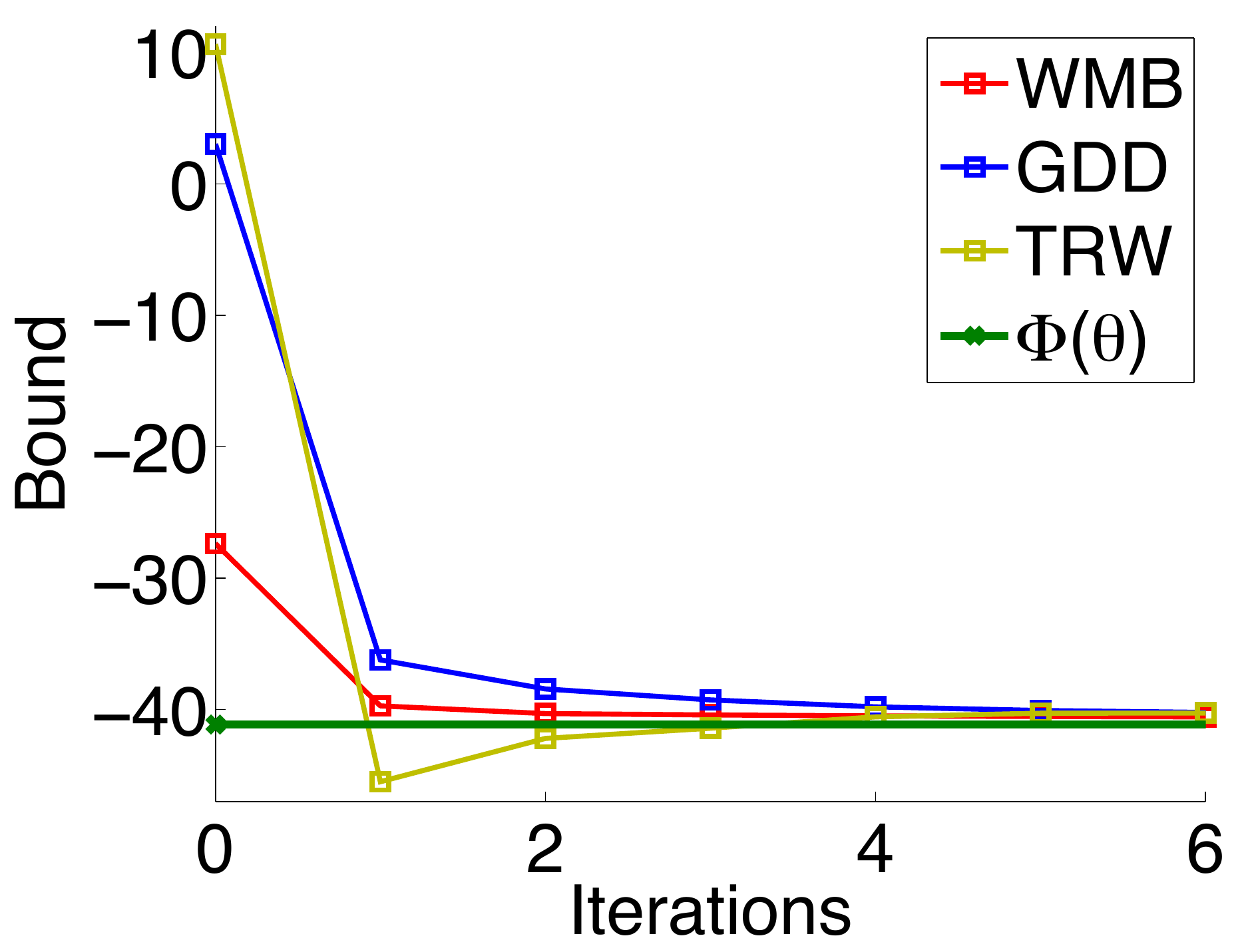} &~~~~~~~~~~~~
\includegraphics[width=4.5cm, clip]{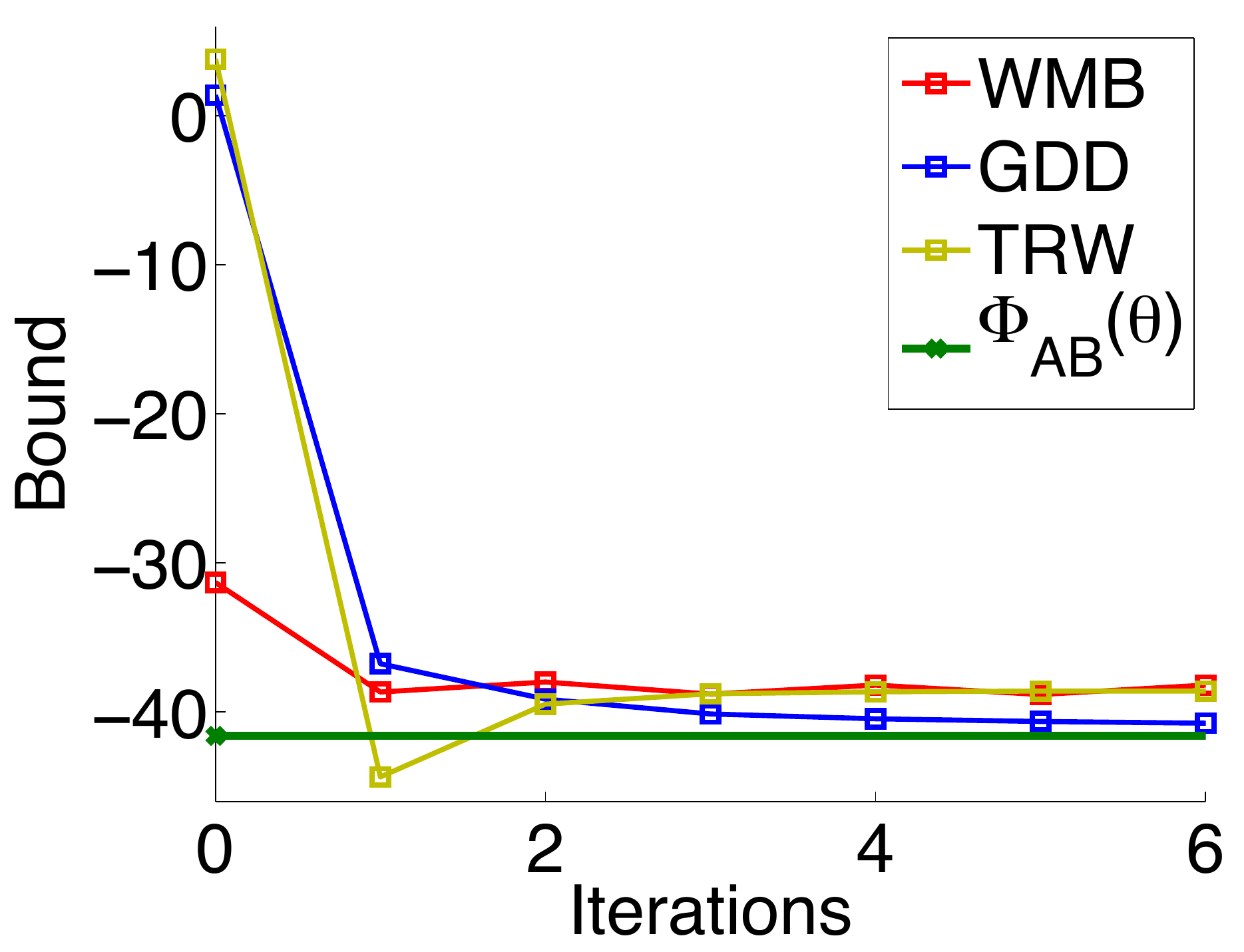}  \\
{ \small (a) sum-inference} 
& {\small (b) marginal MAP} \\
\end{tabular}
\vskip -0.015in
\caption{ Sum-inference and marginal MAP results on a toy Ising model ($5 \times 5$ grid). Each iteration of the different algorithms corresponds to a full sweep over the graph. 
Note that the dual formulation (TRW) is not a bound until convergence; for example, at iteration 1, its objective function
is below the true $\Phi$.
}
\vskip -0.1in
\label{fig:TRBP_dual} %
\end{figure*}

\vspace{-0.5em}
Our GDD directly optimizes a primal bound, and is thus guaranteed to be an upper bound 
of the partition function even before the algorithm converges, enabling a desirable ``any-time" property. 
In contrast, typical implementations of tree reweighted (TRW) belief propagation optimize the dual free 
energy function \citep{wainwright05}, and are not guaranteed to be a bound before convergence. 
We illustrate this point using an experiment on a toy $5\times 5$ Ising grid, with 
parameters generated by normal ditribution $N(0, 2)$ and half nodes selected as max-nodes for marginal MAP.  
Figure~\ref{fig:TRBP_dual}(a)-(b) shows the TRW free energy objective and GDD, WMB upper bounds across iterations; 
we can see that TRW does violate the upper bound property before convergence, while GDD and WMB always give valid upper bounds.  

\vspace{-0.6em}
\section{More Results on Diagnostic Bayesian Networks}
\label{sec:diagnostic_BN}
\begin{figure*}[tb] \centering
\begin{tabular}{cc}
\includegraphics[width=5cm, clip]{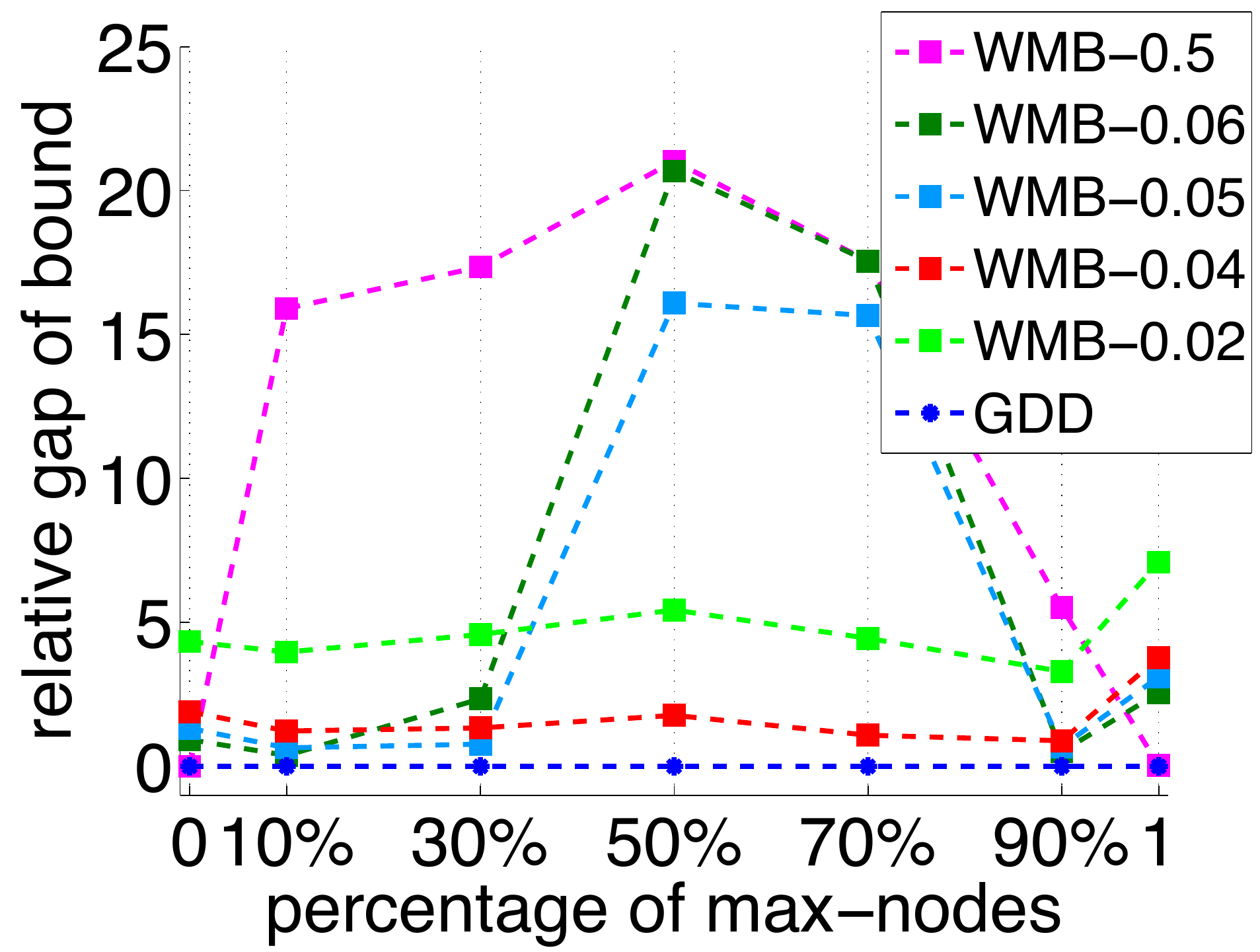} &~~~~~~~~~~~~~~~~~
\includegraphics[width=5cm, clip]{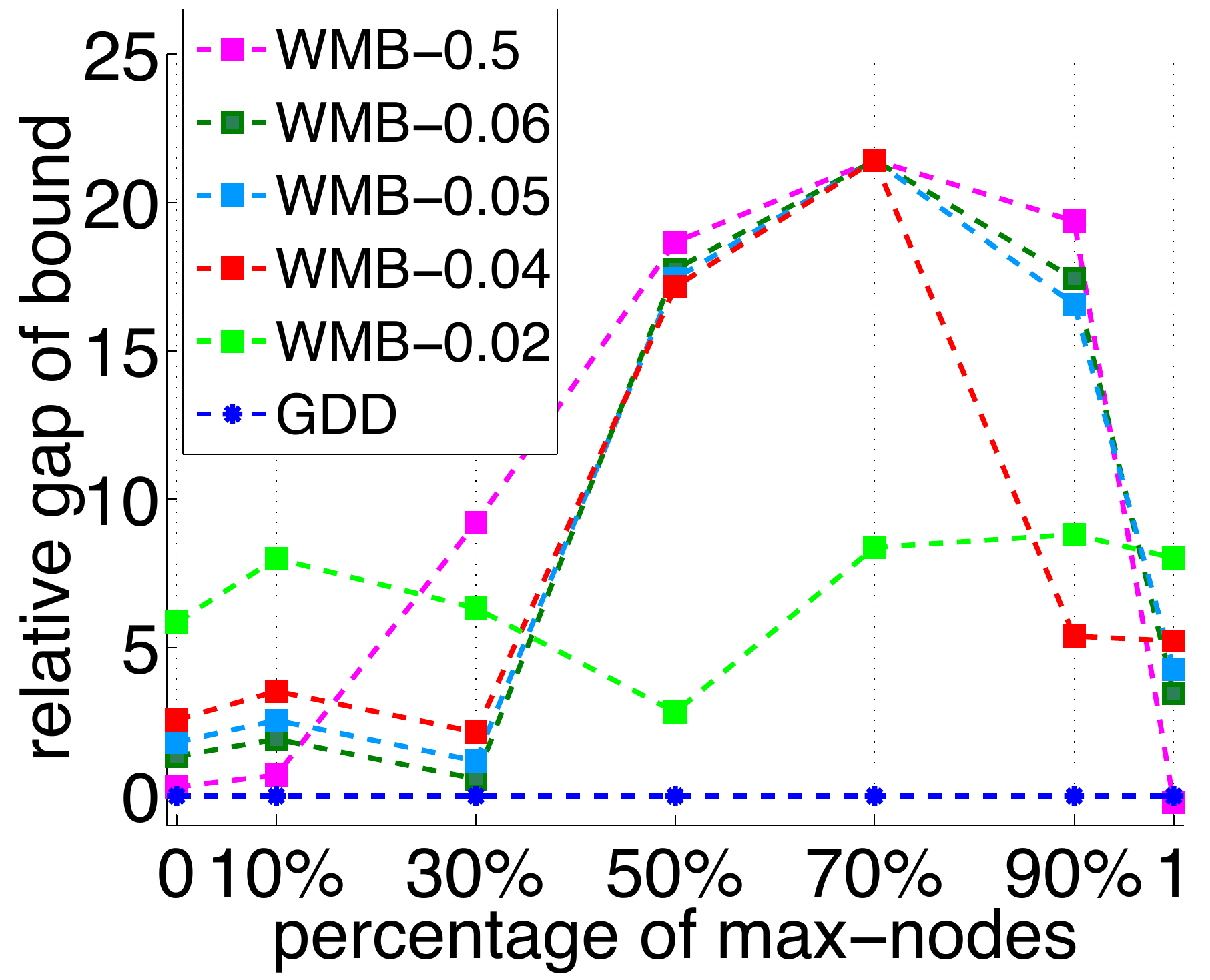} 
 \\
 \vspace{-0.5em}
  {\small (a) BN-1}  &~~~~~~~~~~~~~~~~~  
  {\small (b) BN-2} 
\end{tabular}
\caption{ More marginal MAP results (including sum-inference and MAP) on two diagnostic Bayesian networks.
We report the best results obtained by GDD and WMB with 20 iterations in marginal MAP problems constructed by randomly selecting different percentages of max-nodes. }
\label{fig:moreDN} %
\vspace{0.2em}
\end{figure*}

\vspace{-0.6em}
In addtion to the marginal MAP results on BN-1 and BN-2 in main text,
we vary the percentage of max-nodes when generating the marginal MAP problems; the reported results in Figure~\ref{fig:moreDN}(a)-(b)  are the best bound obtained by the different algorithms with the first 20 iterations. 
In all cases, GDD's results are as good or better than WMB.
WMB-0.5 (WMB with damping ratio $0.5$) appears to work well on sum-only and max-only (MAP) problems, i.e.,
when the percentage of max-nodes equals 0\% and 100\% respectively, but performs very poorly on intermediate settings.
The far more heavily damped WMB-0.04 or WMB-0.02 work better on average, but have much slower convergence. 

\vspace{-0.5em}
\section{More Results on Pedigree Linkage Analysis}
\label{sec:pedigree}
\begin{figure*}[tb] \centering
\begin{tabular}{ccc}
\includegraphics[width=4.30cm, clip]{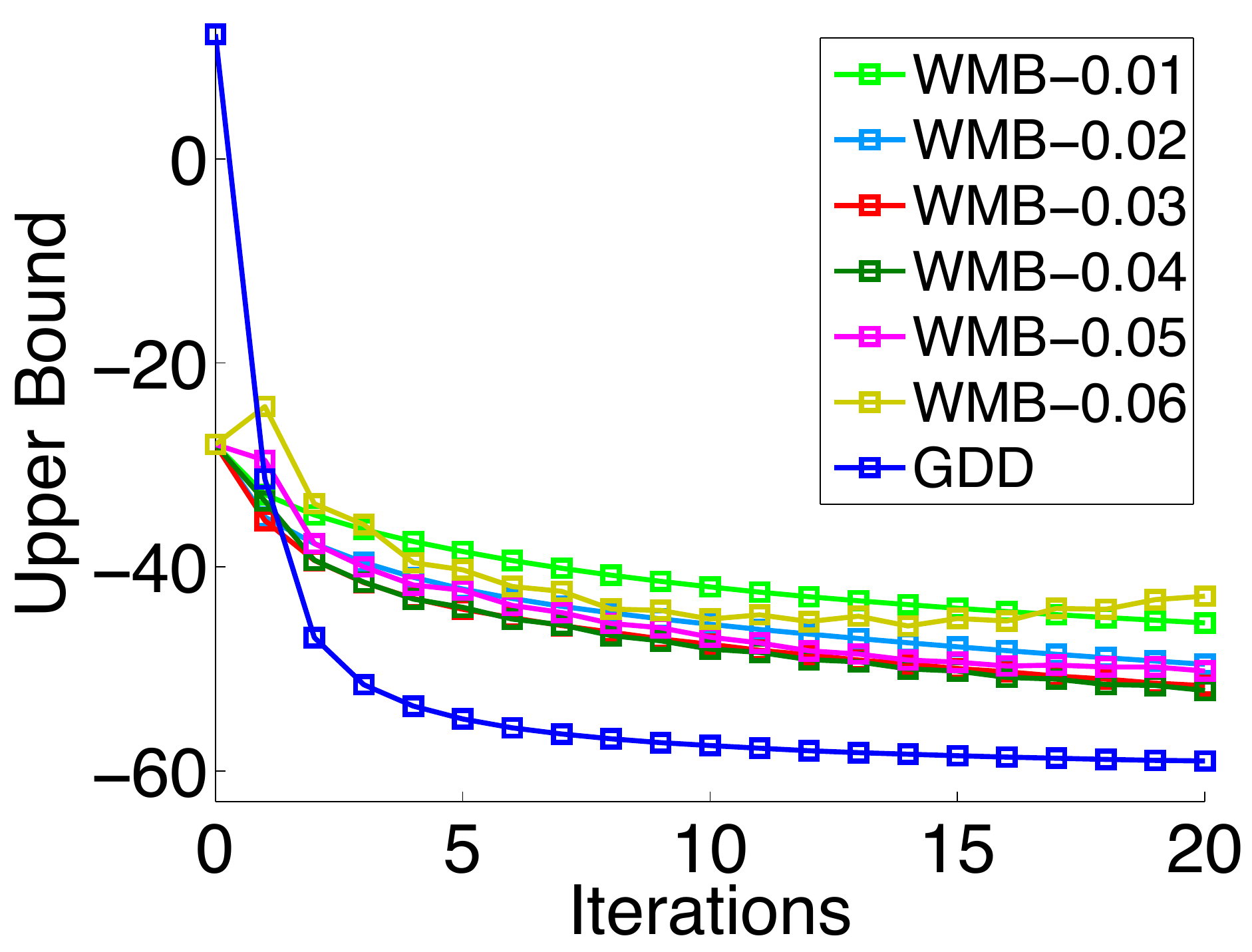} &
\includegraphics[width=4.30cm, clip]{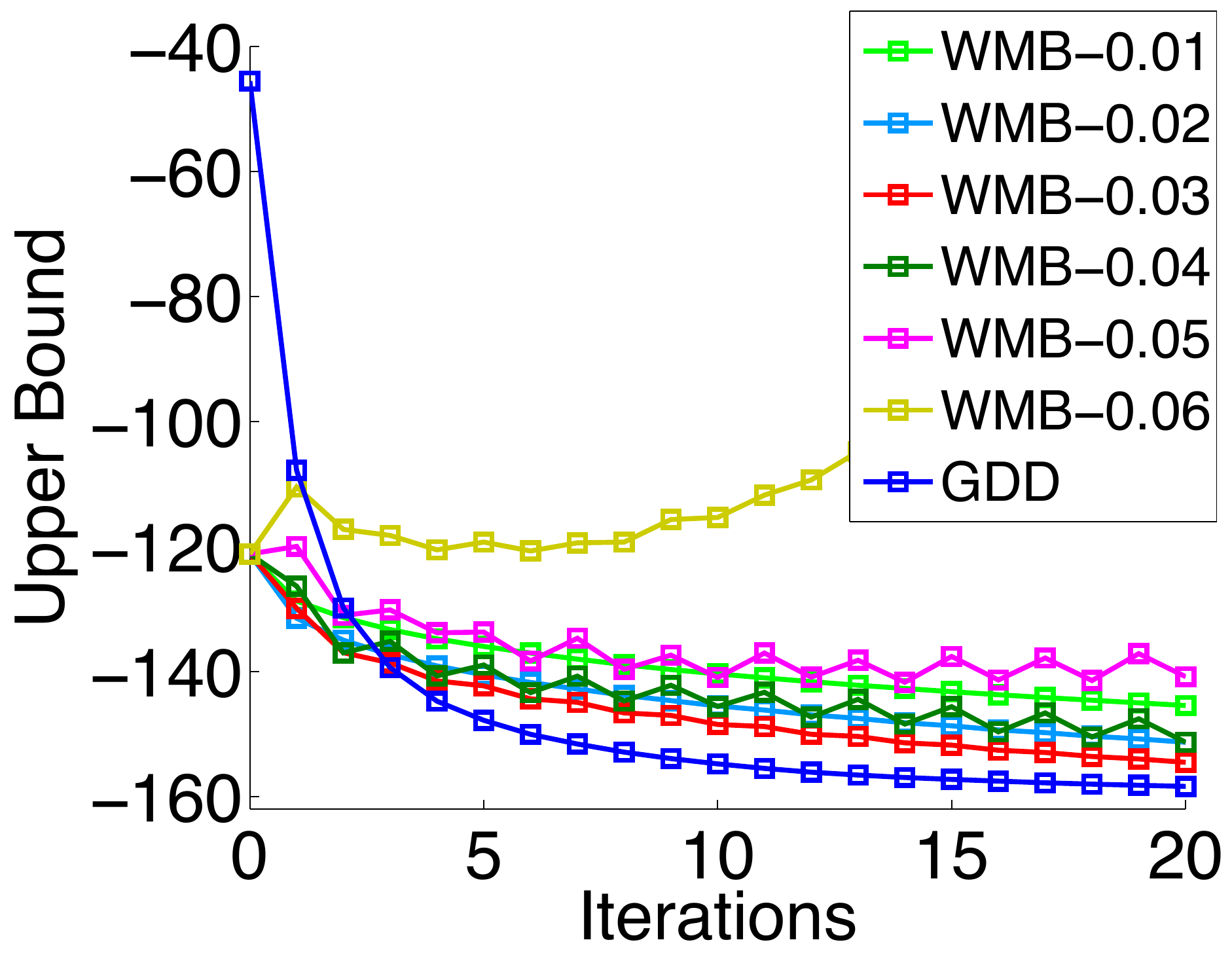} &
\includegraphics[width=4.30cm, clip]{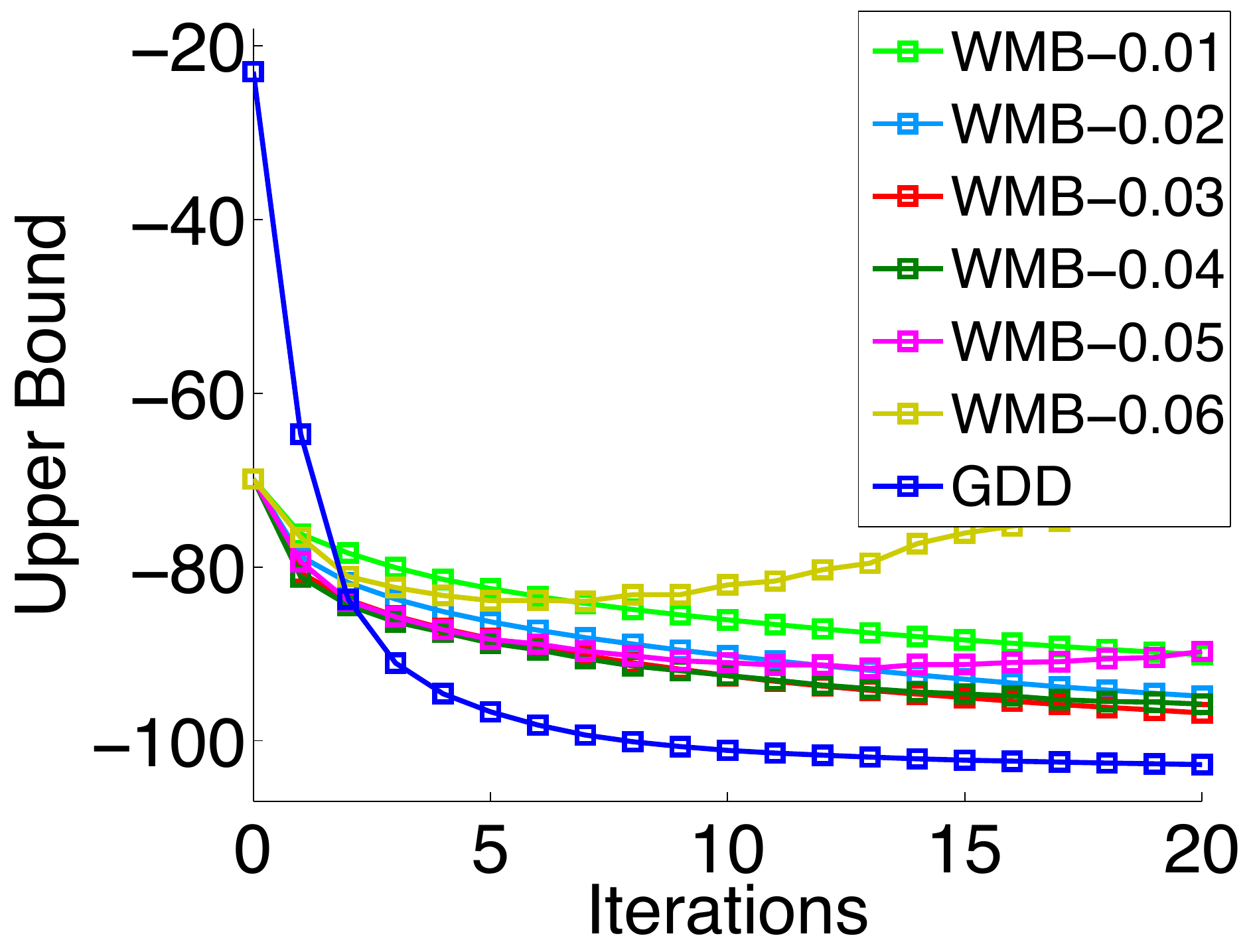}  \\
\vspace{1.2em}
{ \small (a) pedigree13 (1077 nodes) } & {\small (b) pedigree18 (1184 nodes)} & {\small (c) pedigree19 (793 nodes) } 
\\
\includegraphics[width=4.30cm, clip]{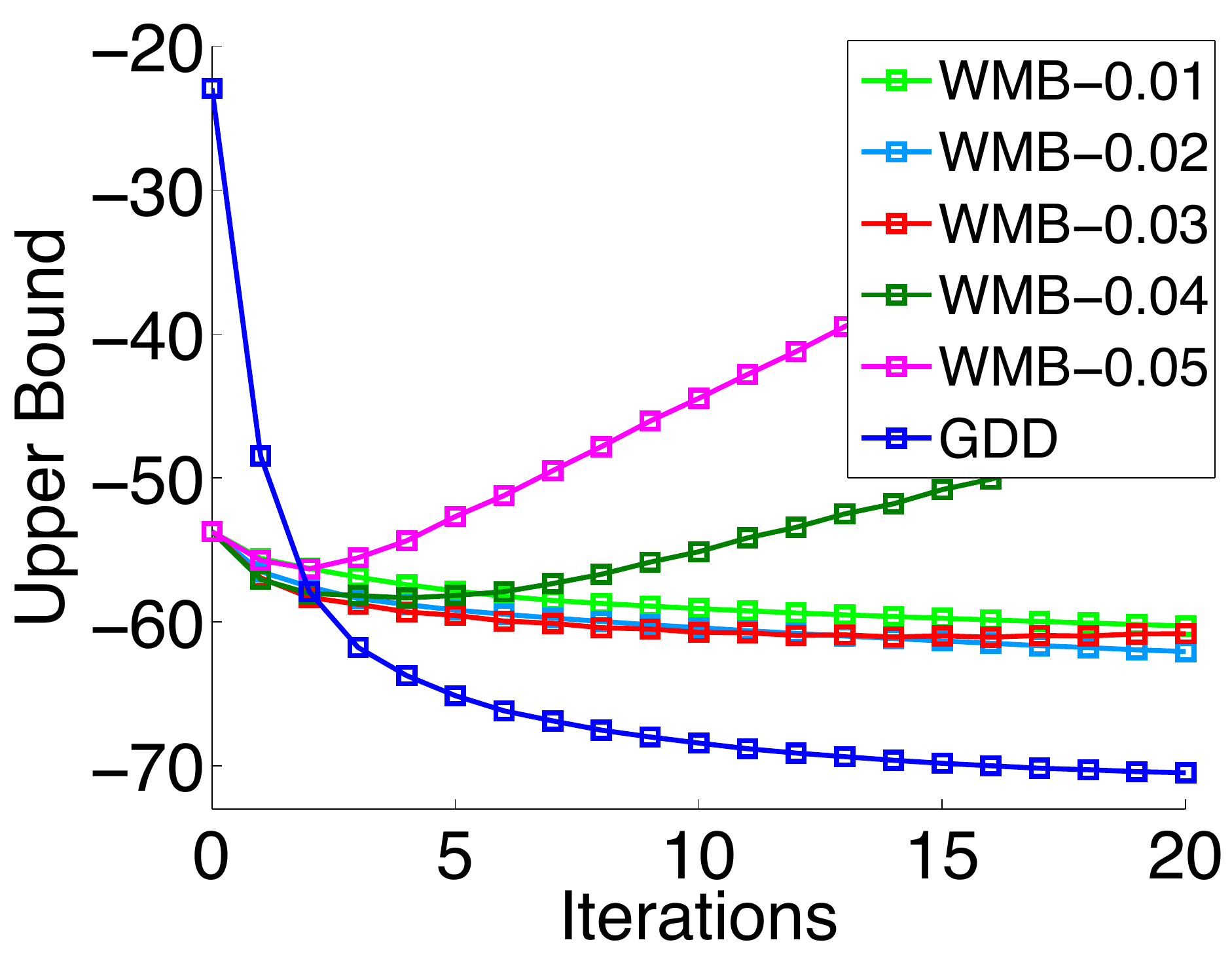} &
\includegraphics[width=4.30cm, clip]{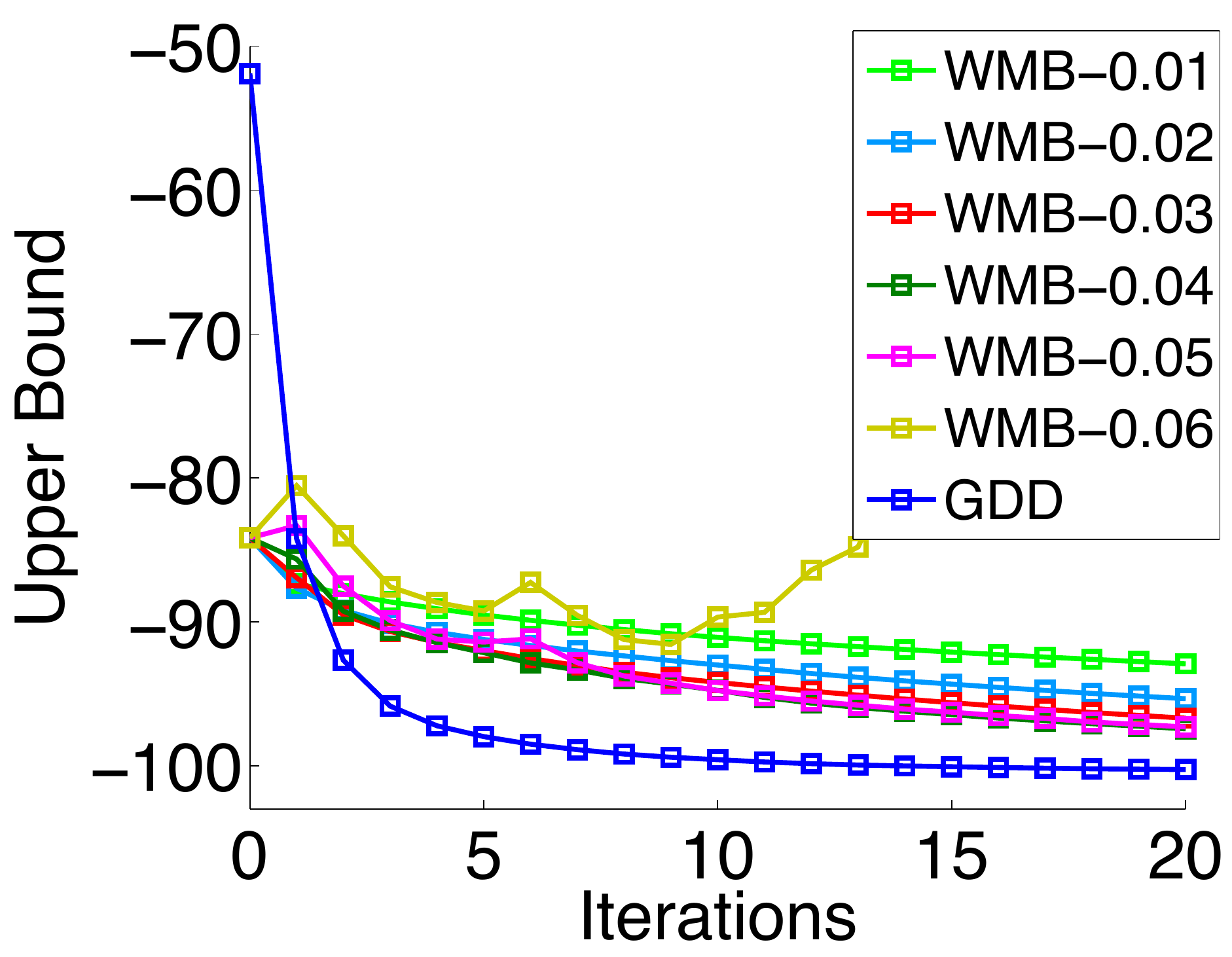} &
\includegraphics[width=4.30cm, clip]{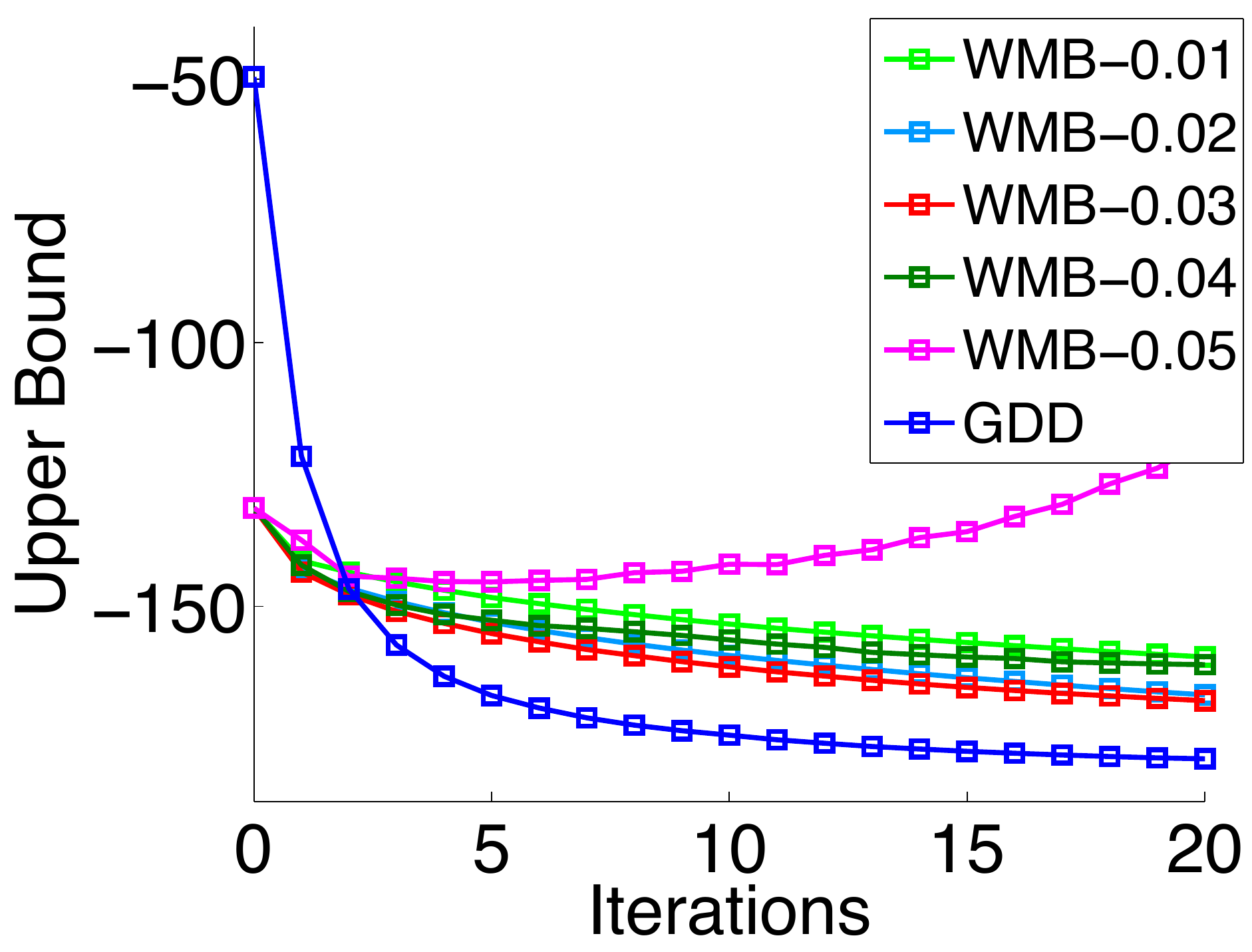}  \\
{ \small (d) pedigree20 (437 nodes) } & {\small (e) pedigree23 (402 nodes)}  & {\small (f) pedigree30 (1289 nodes)} 
\end{tabular}
\vskip -0.01in
\caption{ Marginal MAP inference on additional pedigree linkage analysis models. 
We randomly selected $50\%$ of nodes as max-nodes in these models. 
We tune the damping rate of WMB from 0.01 to 0.06, but we omit WMB-0.06 in the plot if WMB-0.05 is already diverged. }
\label{fig:moreLink_margMAP} %
\end{figure*}

\vspace{-0.6em}
We test our algorithm on additional 6 models of pedigree linkage analysis from the UAI’08 inference challenge.
We construct marginal MAP problems by randomly selected $50\%$ of nodes to be max-nodes, and report all the results in Figure \ref{fig:moreLink_margMAP}. We find that our algorithm consistently outperforms WMB with the best possible damping ratio.

\vspace{-0.5em}
\section{Extensions to Junction Graph}
\label{sec:extension_junction}
\vspace{-0.6em}
Our bound \eqref{bound_wsum}  in the main text  
uses a standard ``factor graph" representation in which the 
cost-shifts $\{\delta_i^{\alpha}\}$ are defined for each variable-factor pair $(i,\alpha)$, 
and are functions of single variables $x_i$.
We can extend our bound to use more general shifting parameters using a junction graph representation;
this allows us to exploit higher order clique structures, leading to better performance. 

Let   $(\mathcal{C}, \mathcal{S} )$ be a junction graph of $p(x; \theta)$ where $\mathcal{C} = \{c\ |\ c \subset V\}$ is the set of clusters, and $\mathcal{S} = \{ s = c_k \cap c_l\ |\  \ c_k, c_l \in \mathcal{C}\}$ is the set of separators. Assume $p(x ; \theta)$ can be reparameterized into the form,
\begin{align}
\label{junction-graph-model}
p(x; \theta) = \exp \Big[ \sum_{c \in \mathcal{C} } \theta_{c} (x_{c}) - \Phi(\theta) \Big],
\end{align}
and the weighted log partition function is rewritten as
$
\Phi_{\vv \w}(\theta) = \log \wsum_{x}^{\vv \w} \exp \Big[  \sum_{c \in \mathcal{C} } \theta_{c} (x_{c}) \Big].
$
Similar to the derivation of bound \eqref{bound_wsum} in the main text, 
we can apply Theorem~\ref{thm:main}, 
 but with a set of more general cost-shifting variables $\delta_{s}^{c}$, defined on each adjacent separator-cluster pair $(s, c)$; this gives the more general upper bound, 
\begin{align}
\label{bound_wsum_junction_graph}
\Phi_{\vv \w} (\theta)  \le 
\sum_{s \in \mathcal{S} } \log\wsum_{ x_s }^{ {\bf w}^{s} } \exp \Big[   \sum_{c \supseteq s } \delta_{s }^{c} (x_{s})  \Big]
 + \sum_{c \in \mathcal{C}} \log\wsum_{x_{c}}^{ {\bf w}^{c} } \exp 
\Big[\theta_{c} (x_{c}) - \sum_{s \subseteq c} \delta_{s}^{c} (x_{s}) \Big],
\end{align}
where we introduce the set of non-negative weights 
${\bf w}^s = \{ w_i^s ~|~ i \in s \}$ on each separator and 
${\bf w}^c = \{ w_i^c ~|~ i \in c \} $ on each cluster,
which should satisfy 
$
 \sum_{s \in N_i^{se} } w_i^{s} + \sum_{c \in N_i^c} w_i^{c} = \w_i,
$
where $N_i^{se} = \{ s ~|~ i \in s \}$ are all the separators that include node $i$, 
and $N_i^c = \{ c ~|~ i \in c \}$ are all the clusters that include node $i$.
Obviously, our earlier bound~\eqref{bound_wsum} in the main text 
can be viewed as a special case of~\eqref{bound_wsum_junction_graph} with a special junction graph whose separators consist of only single variables, that is, $\mathcal{S} = V$. 

A block coordinate descent algorithm similar to Algorithm~\ref{alg:GDD} 
 can be derived to optimize the junction graph bound. In this case, we sweep through all the separators $s$ and perform block coordinate update on all $\{ \delta_{s}^{c}|  \forall  c  \supseteq s\}$ at each iteration. Similarly to Algorithm~1, 
 we can derive a close form update for separators with all-zero weights (that is, $\w_{i} =0$, $\forall i\in s$, corresponding to $s \subseteq B$ in marginal MAP), and perform local gradient descent otherwise.

\vspace{-0.5em}
\section{Proof of Thereom 4.1}
\label{sec:proof_4.1}
\vspace{-0.5em}
\begin{proof}
Note the H\"older's inequality is
$$
\big[ \sum_{x}\prod_j f_j(x)^{1/\xi_0} \big]^{\xi_0} \leq  \prod_j \big [\sum_{x} f_j(x)^{1/\xi_j}\big ]^{\xi_j}, 
$$
where $\{f_j(x)\}$ are arbitrary positive functions, and $\{\xi_j\}$ are non-negative numbers that satisfy $\sum_{j} \xi_j = \xi_0$. 
Note we extend the inequality by defining power sum with $\xi_j =0$ to equal the max operator.
Our result follows by applying  H{\"o}lder's inequality on each $x_i$ sequentially along the elimination order $[x_1, x_2, \cdots, x_n]$.
\end{proof}

\vspace{-0.6em}
\section{Dual Representations}
\label{sec:dual_repres}
\vspace{-0.6em}
\subsection{Background}
\vspace{-0.3em}
The log-partition function $\Phi(\theta)$ has the following variational (dual) form
$$
\Phi(\theta) = \log \sum_{x} \exp(\theta(x)) 
=  \max_{ b \in \mathbb{M}(G) }   \big \{ \langle \theta, b \rangle + H(x; b)  \big \}
$$
where $\mathbb{M}(G)$ is the \emph{marginal polytope} \citep{wainwright2008graphical}.
%
Then, for any scalar $\varepsilon>0$~(including $\varepsilon \rightarrow 0^+$), we have
$$
\Phi_{\varepsilon} (\theta) 
= \varepsilon \log \sum_{x} \exp (\frac{\theta(x)} {\varepsilon} ) 
=  \varepsilon  \max_{ b \in \mathbb{M}}   \big \{ \langle \frac{\theta}{\varepsilon}, b \rangle + H(x; b)  \big \}
=  \max_{ b \in \mathbb{M}}   \big \{ \langle \theta, b \rangle + \varepsilon H(x; b)  \big \}.
$$
As stated in \citep{liu14, liu11}, we can further generalize the variational form of above scalar-weighted log partition function to the vector-{weighted log partition function}~\eqref{powered-sum-inference}  in the main text, 
\begin{align}
\label{exact_variational_form_wsum}
\Phi_{\vv \w} (\theta) 
 = \log \wsum_{x_n}^{\w_n}  \dots\wsum_{x_1}^{\w_1} \exp (\theta(x))
= \max_{ b \in \mathbb{M} (G) }   \big \{ \langle \theta, b \rangle + \sum_{i} \w_i H(x_i | x_{i+1 : n}; b) \big \},
\end{align}
where $H(x_i | x_{i+1 : n}; b)$ is the conditional entropy on $b(x)$, and is defined as
$
H(x_i | x_{i+1 : n}; b) = - \sum_x b(x) \log(b(x_i | x_{i+1 : n} )).
$
See more details of its derivation in Theorem 4.1 within \citep{liu14}.

A notable special case of~\eqref{exact_variational_form_wsum} is the dual form of marginal MAP
\begin{align}
\label{exact_variational_form_marginal_map}
\Phi_{AB}(\theta)
= \max_{x_B}  \log \sum_{x_A} \exp \big(  \theta(x)   \big)
=\max_{ b \in \mathbb{M} (G) }   \big \{ \langle \theta, b \rangle +  H(x_A | x_{B}; b)  \big\},
\end{align}
by setting weights ${\vv \w}_A=1$ and ${\vv \w}_B=0$.

\subsection{Proof of Thereom 4.2}
\label{subsec:proof_4.2}
\vspace{-0.3em}
We now prove the following dual representation of our bound,
\begin{align}
\label{dual_form_GDD_supl}
  \min_{ \delta } L( \delta , {\bf w} ) =
\max_{ {\bf b} \in \mathbb{L}(G) }  \big\{  
	\langle \theta, b \rangle + \sum_{i \in V} w_i H(x_i ; b_i)
	+ \sum_{\alpha \in \mathcal{F} } \sum_{i \in \alpha}  w^{\alpha}_i H(x_i | x_{{\pa}^{\alpha}_i} ; b_{\alpha}) 
	\big\}, 
\end{align}
where $\mathbb{L}(G) = \{ {\bf b} ~|~b_i(x_i)= \sum_{x_{\alpha \setminus i}} b_{\alpha}(x_{\alpha}), ~ \sum_{x_i} b_i(x_i)=1 \}$ is the  local consistency polytope,  and $\pa_i^\alpha = \{ j\in \alpha | j \succ i \}$.
Thereom 4.2 follows directly from \eqref{dual_form_GDD_supl}. 
\begin{proof}
In our primal bound $L( \delta , {\bf w})$~\eqref{bound_wsum}  in the main text, 
we let $\widetilde{\theta_i}(x_i) = \theta_i(x_i) + \sum_{\alpha \in N_i} \delta_i^\alpha (x_i)$
(we add dummy singleton $\theta_i(x_i) \equiv 0 $), \ and 
$\widetilde{\theta}_\alpha (x_\alpha) = \theta_\alpha (x_\alpha) - \sum_{i \in \alpha} \delta_i^\alpha (x_i)$,
then the bound can be rewritten as,
\begin{align*}
L( \widetilde{\theta} , {\bf w}) 
&= \sum_{i\in V} \log \wsum_{x_i}^{w_i} \exp \big[ \widetilde{\theta_i}(x_i) \big]
+ \sum_{\alpha \in \mathcal{F} } \log\wsum_{x_\alpha}^{\bf w^\alpha } 
			\exp \big[  \widetilde{\theta}_\alpha (x_\alpha) \big].
\end{align*}
Note, for any assignment $x$, we have 
$
\sum_i \widetilde{\theta_i} (x_i) + \sum_\alpha \widetilde{\theta}_\alpha (x_\alpha)
= \sum_{\alpha} \theta_\alpha (x_\alpha).
$

By applying the dual form of the powered sum \eqref{exact_variational_form_wsum} on each node and clique respectively, we have
{\small
\begin{align*}
L( \widetilde{\theta} , {\bf w})  
= \sum_{i\in V}   \max_{ b_i \in \mathbb{M}(G_i) }  
			\big\{  \langle \widetilde{\theta_i}, b_i \rangle + w_i H(x_i; b_i)  \big\}
	  + \sum_{\alpha \in \mathcal{F} }  \max_{ b_\alpha \in \mathbb{M}(G_\alpha) }   
	  		\big\{  \langle  \widetilde{\theta}_\alpha, b_\alpha  \rangle   
	  					+ \sum_{i \in \alpha}  w^{\alpha}_i H(x_i | x_{{\pa}^{\alpha}_i} ; b_{\alpha})   \big\} , 
\end{align*} 
}
\!\!\!\!
where ${\pa}^{\alpha}_i$ is the set of variables in $\alpha$ that are summed out later than $i$, 
$\mathbb{M}(G_i)$ and $\mathbb{M}(G_\alpha)$ are the marginal polytopes on singleton node $i$ and clique $\alpha$ respectively,
which enforce $\{ b_i, b_\alpha \}$ to be properly normalized.\\
The above equation can be more compactly rewritten as,
\begin{align*}
L( \widetilde{\theta} , {\bf w})  
=  \max_{ {\bf b} \in \widetilde{\mathbb{M}}  } \Big\{
	  \langle \widetilde{\theta}, b \rangle  
	   + \sum_{i\in V}   w_i H(x_i; b_i) 
	   + 	 \sum_{\alpha \in \mathcal{F} } \sum_{i \in \alpha}  w^{\alpha}_i H(x_i | x_{{\pa}^{\alpha}_i} ; b_{\alpha}) 
	 \Big\},
\end{align*}
where $\widetilde{\mathbb{M}} = \{ \mathbb{M}(G_i), \mathbb{M}(G_\alpha) ~|~ \forall~ i\in V, \alpha \in \mathcal{F} \}$,
and the elements $\{b_i, b_\alpha\}$ of ${\bf b}$ are independently optimized.

Then, by tightening \emph{reparameterization} $\widetilde{\theta} = \{ \widetilde{\theta}_i, \widetilde{\theta}_\alpha \}$,
we have
\begin{align*}
\min_ {\widetilde{\theta} } L( \widetilde{\theta} , {\bf w})  
&= \max_{ {\bf b} \in \widetilde{\mathbb{M}}  }  \min_ {\widetilde{\theta} }  
\Big\{
	 \langle \widetilde{\theta}, b \rangle  
	   + \sum_{i\in V}   w_i H(x_i; b_i) 
	   + 	 \sum_{\alpha \in \mathcal{F} } \sum_{i \in \alpha}  w^{\alpha}_i H(x_i | x_{{\pa}^{\alpha}_i} ; b_{\alpha}) 
\Big\}	  					
\end{align*}
where the order of $\min$ and $\max$ are commuted according to the strong duality~(it's convex with $\widetilde{\theta}$, 
and concave with ${\bf b}$). \\
The inner minimization $\min_ {\widetilde{\theta} }  \langle \widetilde{\theta}, b \rangle  $ is a linear program, 
and it turns out can be solved analytically.
To see this, given the relationship between $\widetilde{\theta} $ and $\delta$, we rewrite the linear program as
\begin{align*}
\min_ {\widetilde{\theta} }  \langle \widetilde{\theta}, b \rangle 
&= \min_{\delta}  \Big\{  
  \langle \theta, b \rangle 
	+ \sum_{i \in V} \sum_{x_i}  \sum_{\alpha \in N_i } \delta_i^\alpha (x_i)  b_i(x_i)
	- \sum_{\alpha \in \mathcal{F} }  \sum_{ x_\alpha }  \sum_{i \in \alpha} \delta_i^\alpha(x_i)   b_\alpha(x_\alpha) 
\Big\}, \\
&=  \min_{\delta}  \Big\{ 
	\langle \theta, b \rangle
	+ \sum_{(i, \alpha)}  \sum_{x_i}  \delta_i^\alpha (x_i) \Big( b_i(x_i)   - \sum_{x_{\alpha \backslash i}} b_{\alpha}(x_\alpha)  \Big)
\Big\}.
\end{align*}
Then, it is easy to observe that the linear program is either equal to $\langle \theta, b \rangle$ only if $b$ satisfy the marginalization constraint
$
\sum_{x_{\alpha \backslash i}} b_{\alpha}(x_\alpha) = b_i(x_i)
$
for $\forall (i, \alpha)$, 
or  it will become negative infinity.
Considering the outer maximization, we have
\begin{align*}
 \min_{ \widetilde{\theta} } L( \widetilde{\theta} , {\bf w} ) =
\max_{ {\bf b} \in \mathbb{L}(G) }  \big\{  
	\langle \theta, b \rangle + \sum_{i \in V} w_i H(x_i ; b_i)
	+ \sum_{\alpha \in \mathcal{F} } \sum_{i \in \alpha}  w^{\alpha}_i H(x_i | x_{{\pa}^{\alpha}_i} ; b_{\alpha}) 
	\big\}, 
\end{align*}
where $\mathbb{L}(G)$ is the local consistency polytope that is obtained by enforcing both $\widetilde{\mathbb{M}} $ and the marginalization constraint.
\end{proof}

\subsection{Connection with Existing Free Energy Forms}
\label{subsec:connection_free_energy}
\vspace{-0.3em}
Most variational forms are expresssed in the following linear combination of local entropies 
\citep{yedidia2005constructing, hazan10},
\begin{align}
\label{general_free_energy}
\langle \theta, b \rangle + \sum_{\beta} c_\beta H(b_\beta),
\end{align}
where $\beta$ refers the region, $c_\beta$ refers the general counting number, $b_\beta(x_\beta)$ is the local belief.

We can rewrite our dual representations~\eqref{dual_form_GDD_supl} as,
$$
\langle \theta, b \rangle + \sum_{i \in V} w_i H(x_i ; b_i)
	+ \sum_{\alpha \in \mathcal{F} } \sum_{i \in \alpha}  w_i^\alpha \big ( H(x_i, x_{\pa_i^\alpha} ~;~ b_\alpha ) 
										- H(x_{\pa_i^\alpha} ~;~ b_\alpha)  \big),
$$
where $\pa_i^\alpha$ is the set of variables in $\alpha$ that rank later than $i$. 
Without loss of generality, assuming $x_\alpha = [x_1, \cdots, x_i, x_j, \cdots x_c]$, i.e. $x_i$ and $x_j$ are adjacent in the order,
we can get
\begin{align}
\label{rewrite_dual_of_GDD}
\langle \theta, b \rangle + \sum_{i \in V} w_i H(x_i ; b_i) 
	+ \sum_{\alpha \in \mathcal{F} } 
	\Big\{
		w_1^\alpha H(x_\alpha; b_\alpha) 
		+ \sum_{[i,j] \sqsubseteq \alpha} ( w_j^\alpha - w_i^\alpha) H(x_{\pa_i^\alpha}~;~ b_{\pa_i^\alpha} ) 
	\Big\}
\end{align}
where belief $b_{\pa_i^\alpha}$ is defined by 
$b_{\pa_i^\alpha} (x_{\pa_i^\alpha}) = \sum_{x_{\alpha \backslash \pa_i^\alpha} } b_\alpha (x_\alpha)$.

One can view \eqref{rewrite_dual_of_GDD} in terms of \eqref{general_free_energy}, by selecting the region 
$\beta \in \{ i \in V \}  \cup  \{\alpha \in \mathcal{F} \} \cup \{  \pa_i^\alpha ~|~ \forall (i, \alpha) \}$;
some counting numbers $c_\beta$ will be the differences of weights $ w_j^\alpha - w_i^\alpha$.

\subsection{Matching Our Bound to WMB}
\vspace{-0.3em}
After the weights are optimized, our GDD bound matches to WMB bound with optimum weights.
A simple weight initialization method matches our bound to WMB with uniform weights on each mini-bucket,
which often gives satisfactory result; 
a similar procedure can be used to match the bound with more general weights as in Section~\ref{sec:extension_junction}.
We first  set $w_i = 0$ for all nodes $i$. 
We then visit the nodes $x_i$ along the elimination order ${\bf o} = [x_1, x_2, \cdots, x_n]$,
and divide $x_i$'s neighborhood cliques $N_i = \{ \alpha | \alpha \ni i\}$ into two groups: 
(1) the \emph{children cliques} in which all $x_{\alpha \backslash i}$ have already been eliminated, that is, $N_i^{ch} = \{ \alpha ~|~ \forall j \!\in\! \alpha \backslash i, \ 
j \prec i
\text{\ in \ }  {\bf o} \}$; 
(2) the other, \emph{parent cliques} $N_i^{pa} =  \{ \alpha ~|~ \exists j \in \alpha \backslash i, \ 
j \succ i 
\text{\ in \ }  {\bf o} \}$.
We set $w_i^\alpha = 0$ for all the children cliques ($\alpha \in N_i^{ch}$), and uniformly split the weights, that is, $w_i^\alpha = \w_i / |N_i^{pa}|$, across the parent cliques.

\vspace{-0.5em}
\section{Proof of Therom 5.1}
\label{sec:Proof_Theorem5.1}
\vspace{-0.5em}
\begin{proof}
For each $\delta_i^{\alpha}(x_i)$,  the involved terms in $L(\delta, {\bf w})$ are
$
L_i^\alpha(\delta) = \Phi_{w_i} (\delta) + \Phi_ {\bf w^\alpha} (\delta ), 
$
where 
\begin{align*}
\Phi_{w_i} (\delta)  =  \log\wsum_{x_i}^{w_i} \exp \Big[ \sum_{\alpha \in N_i} \delta_i^{\alpha}(x_i) \Big],
\quad%
\Phi_ {\bf w^\alpha} (\delta ) =  \log\wsum_{x_\alpha}^{\bf w^\alpha } \exp \Big[\theta_{\alpha} (x_\alpha) -\sum_{i\in \alpha} \delta_i^{\alpha}(x_i) \Big]. 
\end{align*}
Our result follows by showing that 
\begin{align*}
&\frac{ \partial \Phi_{w_i} (\delta) }  { \partial \delta_i^{\alpha} (x_i) } 
= \mu_i (x_i)
 \quad \quad \quad \quad \ \ \text{and} \quad
\frac{ \partial \Phi_{w_i} (\delta) } { \partial w_i }  
= H(x_i; \mu_i), \\
&\frac{ \partial \Phi_ {\bf w^\alpha} (\delta ) } { \partial \delta_i^{\alpha} (x_i) }  
= -  \sum_{ x_{\alpha \backslash i} } \mu_{\alpha}(x_{\alpha})
 \quad \text{and} \quad
 \!\!
\frac{ \partial \Phi_ {\bf w^\alpha} (\delta ) } { \partial w_i^\alpha }  
= H(x_i | x_{i+1 : c} ; \mu_\alpha ).
\end{align*}
The gradient of $\Phi_{w_i} (\delta)$ is straightforward to calculate, 
\begin{align*}
\frac { \partial \Phi_{w_i}   } { \partial \delta_i^\alpha (x_i) } 
 = \frac{\partial}{\partial \delta_i^\alpha (x_i)} 
 \Big( w_i \log \sum_{x_i}  \exp  \Big[ \frac{\sum_{\alpha \in N_i} \delta_i^{\alpha}(x_i)} {w_i} \Big]  \Big)
 = \frac{  \exp  \big[ \frac{\sum_{\alpha \in N_i} \delta_i^{\alpha}(x_i)} {w_i} \big]   } 
 { Z_{w_i}  }
 = \mu_i (x_i),
\end{align*}
where $Z_{w_i} =  \sum_{x_i} \exp  \big[ \frac{\sum_{\alpha \in N_i} \delta_i^{\alpha}(x_i)} {w_i} \big]$, and
\begin{align*}
\!\!\!\!\!\!\!\!\!\!\!\!\!\!\!\!\!\!\!\!\!\!
\frac { \partial \Phi_{w_i}   } { \partial w_i }  
&= \log Z_{w_i}  + w_i  \cdot  \frac{1}{Z_{w_i}} 
	 \cdot  
	\sum_{x_i} \Big\{ \exp \Big[ \frac{ \sum_{\alpha \in N_i} \delta_i^{\alpha}(x_i)  }   {w_i} \Big]  
				\cdot  \frac{  \sum_{\alpha \in N_i} \delta_i^{\alpha}(x_i)  }  {-w_i^2} \Big\}  \\
&= \log Z_{w_i}  -  \sum_{x_i} \Big\{  \mu_i (x_i)  \cdot  \frac{ \sum_{\alpha \in N_i} \delta_i^{\alpha}(x_i) }  {w_i}  \Big\}   \\
&= -  \sum_{x_i} \Big\{  \mu_i (x_i)  \cdot  \Big[ \frac{ \sum_{\alpha \in N_i} \delta_i^{\alpha}(x_i) }  {w_i}  -  \log Z_{w_i}  \Big]  \Big\}
= H(x_i ; \mu_i).
\end{align*}
The gradient of $\Phi_ {\bf w^\alpha} (\delta )$ is more involved; see Proposition~\ref{pro_clique_grdt} for a detailed derivation.

Given the gradients, the moment matching condition~\eqref{moment_matching} in Therom~\ref{thm:matching} obviously holds.
We now prove the entropy matching condition in \eqref{entropy_matching}. 
The constraint optimization is
$$
\min_{\bf w} L ( {\bf w}), \quad
\text{ s.t.  } 
w_i \ge 0, \  w_i^\alpha \ge 0, \  w_i + \sum_\alpha w_i^\alpha = \w_i .
$$
Note, when $\w_i =0$, the optimization is trival, so we simply assume $\w_i > 0$. 
We frame the Lagrangian as
$$
 K ({\bf w}, \vv \lambda, \vv g)  \eqdef
 L( {\bf w}) + \sum_i \lambda_i \big(  w_i + + \sum_\alpha w_i^\alpha - \w_i \big) 
+ \sum_i g_i w_i + \sum_{(i, \alpha)} g_i^\alpha w_i^\alpha .
$$
Note $\vv g \le 0$~(dual feasibility), otherwise $\max_{\vv g, \vv \lambda} K ({\bf w}, \vv \lambda, \vv g) $ will approach infinity.
The KKT conditions are
\begin{align}
\text{stationarity: \quad}
\label{stationary_wi}
&\frac { \partial K   } { \partial w_i }  
= H(x_i; \mu_i)  + \lambda_i  + g_i = 0,\\
\quad
\label{stationary_wi_alpha}
&\frac { \partial K   } { \partial w_i^\alpha }  
= H(x_i | x_{i+1 : c} ; \mu_\alpha ) + \lambda_i + g_i^\alpha  = 0,    
\\
\label{complementary slackness}
\text{complementary slackness: \quad} 
&g_i w_i = 0,
\quad 
g_i^\alpha w_i^\alpha = 0.  
\end{align}
We multiply $w_i$ and $w_i^\alpha$ to \eqref{stationary_wi} and \eqref{stationary_wi_alpha} respectively,
then we can eliminate the KKT multipliers $g_i$ and $g_i^\alpha$ by applying the complementary slackness~\eqref{complementary slackness}, 
\begin{align}
\label{st_cs_together_i}
w_i H(x_i; \mu_i) + w_i \lambda_i = 0, \\
\label{st_cs_together_i_alpha}
w_i^\alpha H(x_i | x_{i+1 : c} ; \mu_\alpha ) + w_i^\alpha \lambda_i = 0.
\end{align}
By summing \eqref{st_cs_together_i_alpha} over all $\alpha \in N_i$ and adding \eqref{st_cs_together_i},
 we can solve the multiplier $\lambda_i$ as
\begin{align*}
\lambda_i = - \frac{ w_i H(x_i; \mu_i)  +  \sum_{\alpha}  w_i^\alpha H(x_i | x_{i+1 : c} ; \mu_\alpha )   }  {\w_i}
= - \bar{H_i} .
\end{align*}
We plug it into \eqref{st_cs_together_i} and \eqref{st_cs_together_i_alpha}, and obtain the entropy matching condition~\eqref{entropy_matching}
 in the Therom~\ref{thm:matching}.
\end{proof}

\vspace{-0em}
\section{Derivations of Closed-form Update}
\label{sec:Close-form}
\vspace{-0em}
We first derive the closed-form update rule for $\delta_i^\alpha (x_i)$ in Proposition \ref{pro_close-form-update}.
We derive the closed-form update rule for the block $\vv \delta_{N_i} = \{ \delta_i^\alpha (x_i) ~|~ \forall \alpha \in N_i \}$ in Proposition \ref{pro_close-form-star-update}.

\begin{pro}
\label{pro_close-form-update}
Given max node $i$ in marginal MAP (i.e., $\w_i=0$ ) and one clique $\alpha \ni i$~(i.e. $i \in \alpha$), 
keeping all $\delta$ fixed except $\delta_i^{\alpha}(x_i)$, there is a closed-form update rule,
\begin{align}
\delta_i^{\alpha}(x_i) \leftarrow  \frac{1}{2} \log \wsum_{x_{\alpha \backslash i}}^{ w_{\backslash i}^{\alpha} } 
 \exp \big[ \theta_{\alpha}(x_\alpha) - \sum_{j \in \alpha \backslash i} \delta_j^{\alpha}(x_j) \big]
- \frac{1}{2}  \sum_{\beta \in N_i \backslash \alpha}\delta_i^{\beta}(x_i),
\label{close-form-update}
\end{align}
where $x_{\alpha \backslash i}  = \{ x_j :  j\in \alpha, j \neq i \}$,  
$w_{\backslash i}^{\alpha} =  \{ w_j^\alpha : j\in \alpha, j\neq i \}$,  
and $N_i = \{ \alpha | \alpha \ni i \}$ is the set of all clique factors in the neighborhood of node $i$. 
Futhermore, this update will monotonically decrease the upper bound.
\begin{proof} 
The terms within the bound $L(\delta,  {\bf w})$ that depend on $\delta_i^{\alpha}(x_i)$ are,
\begin{align}
& \max_{x_i} \big[  \sum_{\alpha\in N_i} \delta_i^{\alpha}(x_i) \big]
+ \max_{x_i} \log \wsum_{x_{\alpha \backslash i}}^{ w_{\backslash i}^{\alpha} } \exp \big[ \theta_{\alpha}(x_\alpha) 
- \sum_{i \in \alpha} \delta_i^{\alpha}(x_i)  \big] .
\label{before_clique_2_max}
\end{align}
The sub-gradient of (\ref{before_clique_2_max}) w.r.t. $\delta_i^{\alpha}(x_i)$ equal to zero if and only if,
\begin{align*}
x_i^* = \argmax_{x_i} \big[  \sum_{\alpha\in N_i} \delta_i^{\alpha}(x_i) \big]
= \argmax_{x_i} \, \log \wsum_{x_{\alpha \backslash i}}^{ w_{\backslash i}^{\alpha} } \exp \big[ \theta_{\alpha}(x_\alpha) 
- \sum_{i \in \alpha} \delta_i^{\alpha}(x_i)  \big] ,
\end{align*}
which is ``$\argmax$'' matching. One sufficient condition of this matching is,
\begin{align*}
 \sum_{\alpha\in N_i} \delta_i^{\alpha}(x_i)
= \log \wsum_{x_{\alpha \backslash i}}^{ w_{\backslash i}^{\alpha} } 
\exp \big[ \theta_{\alpha}(x_\alpha) - \sum_{i \in \alpha} \delta_i^{\alpha}(x_i)  \big] 
\end{align*}
which impllies matching of ``pseudo marginals''. 
Then, one can pull $\delta_i^\alpha(x_i)$ outside from the operator
 $\log \wsum_{x_{\alpha \backslash i}}^{ w_{\backslash i}^{\alpha} }  \exp$, 
and get  the closed-form equation
\begin{align*}
 \delta_i^{\alpha}(x_i)
=  \frac{1}{2}\log \wsum_{x_{\alpha \backslash i}}^{ w_{\backslash i}^{\alpha} } 
 \exp \big[ \theta_{\alpha}(x_\alpha) - \sum_{j \in \alpha \backslash i} \delta_j^{\alpha}(x_j) \big]
   - \frac{1}{2} \sum_{\beta \in N_i \backslash \alpha}\delta_i^{\beta}(x_i) .
\end{align*}

To prove monotonicity,  we substitute above update equation of $ \delta_i^{\alpha}(x_i)$
into (\ref{before_clique_2_max}); then  we get,
\begin{align}
\max_{x_i} \Big\{  \sum_{\beta \in N_i \backslash \alpha }\delta_i^{\beta}(x_i)
+  \log \wsum_{x_{\alpha \backslash i}}^{ w_{\backslash i}^{\alpha} } 
 \exp \big[ \theta_{\alpha}(x_\alpha) - \sum_{j \in \alpha \backslash i} \delta_j^{\alpha}(x_j)  \big]  \Big\} .
\label{after_clique_2_max}
\end{align}
Clearly, (\ref{after_clique_2_max}) $\le$ (\ref{before_clique_2_max})  by using the fact that $\max_x[ f(x) + g(x) ] \le \max_x f(x) + \max_x g(x)$.
\end{proof}
\end{pro}

\begin{pro}
\label{pro_close-form-star-update}
Given node $i\in B$ (i.e., a max node)  and all neighborhood cliques $N_i = \{ \alpha | \alpha \ni i \}$, we can jointly optimize 
$\vv \delta_{N_i} = \{ \delta_i^{\alpha}(x_i) ~|~ \forall \alpha \in N_i\}$ in closed-form 
by keeping the other $\{ \delta_j^\alpha ~|~ j \neq i, \forall \alpha \in N_i  \}$ fixed. 
The update rule is,
\begin{align}
\label{closed-form-clique-star-update}
\delta_i^{\alpha}(x_i) \leftarrow &\ 
\frac{|N_i|}{|N_i| + 1} \gamma_i^{\alpha}(x_i)  
- \frac{1}{|N_i| +1 }  \!\!\!  \sum_{\beta\in N_i \backslash \alpha} \!\!\! \gamma_i^{\beta}(x_i), 
\end{align}
where $|N_i|$ is the number of neighborhood cliques, and 
$\{ \gamma_i^{\alpha}(x_i) ~|~\forall \alpha \in N_i \}$ are defined by
\begin{align}
\label{intermediate_gamma}
\gamma_i^{\alpha}(x_i) = \log \wsum_{x_{\alpha \backslash i} }^{ {\bf w}_{\backslash i}^{\alpha}}  
\exp \big[ \theta_{\alpha}(x_\alpha) - \sum_{j \in \alpha \backslash i } \delta_j^{\alpha}(x_j) \big].
\end{align}

Futhermore, this upate will monotonically decrease the upper bound.
\begin{proof}
For $\forall \alpha \in N_i$, we have closed-form solutions for $\delta_i^{\alpha}(x_i)$ as Proposition \ref{pro_close-form-update}. 
We rewrite it as,
\begin{align}
\label{close_form_update_Eq}
\forall \alpha \in N_i, \quad
2 \delta_i^{\alpha}(x_i) +  \sum_{\beta \in N_i \backslash \alpha}\delta_i^{\beta}(x_i)
=   \log \wsum_{x_{\alpha \backslash i}}^{ w_{\backslash i}^{\alpha} } 
 \exp \big[ \theta_{\alpha}(x_\alpha) - \sum_{j \in \alpha \backslash i} \delta_j^{\alpha}(x_j) \big].
\end{align}
Note, for $\forall \alpha, \beta \in N_i$,  there is a linear relationship between $\delta_i^{\alpha}(x_i)$ and $\delta_i^{\beta}(x_i)$. 

We denote column vector $\vv \gamma_i (x_i)$ filled $\alpha$-th element with
$$
\gamma_i^\alpha (x_i) =  \log \wsum_{x_{\alpha \backslash i}}^{ w_{\backslash i}^{\alpha} } 
 \exp \big[ \theta_{\alpha}(x_\alpha) - \sum_{j \in \alpha \backslash i} \delta_j^{\alpha}(x_j) \big].
$$
We also frame all $\{ \delta_i^{\alpha} (x_i) ~|~ \alpha \in N_i \}$ into a column vector $\vv \delta_{N_i}(x_i) $, 
and denote $|N_i| \times |N_i|$ matrix~A
$$
A = 
\begin{pmatrix}
  2 & 1 & \cdots & 1 \\
  1 & 2 & \cdots & 1 \\
  \vdots  & \vdots  & \ddots & \vdots  \\
  1 & 1 & \cdots & 2
 \end{pmatrix}, 
 \text{ and note }
 A^{-1} =
\begin{pmatrix}
  \frac{|N_i|}{|N_i| + 1} & -\frac{1}{|N_i| + 1} & \cdots & -\frac{1}{|N_i| + 1} \\
  -\frac{1}{|N_i| + 1}  & \frac{|N_i|}{|N_i| + 1} & \cdots & -\frac{1}{|N_i| + 1} \\
  \vdots  & \vdots  & \ddots & \vdots  \\
  -\frac{1}{|N_i| + 1} & -\frac{1}{|N_i| + 1} &  \cdots & \frac{|N_i|}{|N_i| + 1}. 
 \end{pmatrix}
$$
It is easy to verify
$
A {\vv \delta_{N_i} } (x_i) = \vv\gamma_i(x_i) .
$
from \eqref{close_form_update_Eq}. 
 Since $A$ is invertable, one can solve 
$$
\vv \delta_{N_i} (x_i) = A^{-1}  \vv\gamma_i(x_i) .
$$ 
Then, one can read out the closed-form update rule~\eqref{closed-form-clique-star-update}.
The monotonicity holds directly by noticing that the update rule
(\ref{closed-form-clique-star-update}) are solutions which jointly satisfy equation (\ref{close-form-update}).
\end{proof}
\end{pro}

\vspace{-0.5em}
\section{Derivations of Gradient}
\label{sec:supp_gradient}
\vspace{-0.5em}

\begin{pro}
\label{pro_clique_grdt}
Given a weight vector ${\bf w^\alpha} = [w_1^\alpha, \cdots, w_i^\alpha, \cdots, w_c^\alpha]$ 
associated with variables $x_\alpha = \{ x_1, \cdots, x_i, \cdots, x_c \}$ on clique $\alpha$, where $c = |\alpha|$
the power  sum over clique $\alpha$ is,
{ \small
\begin{align*}
\Phi_ {\bf w^\alpha} (\delta )
=\log\wsum_{x_\alpha}^{\bf w^\alpha } \exp \Big[\theta_{\alpha} (x_\alpha) -\sum_{i\in \alpha} \delta_i^{\alpha}(x_i) \Big]
= \log \wsum_{x_c}^{w_c^\alpha} \cdots \wsum_{x_i}^{w_i^\alpha} \cdots \wsum_{x_1}^{w_1^\alpha}
\exp \Big[\theta_{\alpha} (x_\alpha) -\sum_{i\in \alpha} \delta_i^{\alpha}(x_i) \Big].
\end{align*}
}
We recursively denote $Z_i$ as the partial power sum up to $x_{1:i}$, 
\begin{align}
Z_0 (x_\alpha)  = \exp \Big[\theta_{\alpha} (x_\alpha) -\sum_{i\in \alpha} \delta_i^{\alpha}(x_i) \Big]
\quad \text{and} \quad
Z_i (x_{i+1:c}) = \wsum_{x_i}^{w_i^\alpha} Z_{i-1} (x_{i:c}),
\label{recur_Zi}
\end{align}
thus 
$
%
\log Z_c = \Phi_ {\bf w^\alpha} .
$
We also denote the ``pseudo marginal'' (or, belief)  on $x_\alpha$,
\begin{align*}
\mu_\alpha(x_\alpha) 
= \prod_{i=1}^{c} \mu_\alpha (x_i | x_{i+1:c}); \quad
\mu_\alpha(x_i | x_{i+1:c})
 = \Big( \frac { Z_{i-1} (x_{i:c})  } { Z_i (x_{i+1:c})  } \Big) ^{ 1 / w_i^\alpha } ,
\end{align*}
and it is easy to verify that $\mu_\alpha(x_i | x_{i+1:c})$ and $\mu_\alpha(x_\alpha)$ are normalized.

Then, the derivative of $\Phi_{\bf w^\alpha}$ w.r.t. $\delta_i^\alpha (x_i)$ can be written by beliefs,
\begin{align}
\label{grdt_delta_i_alpha}
\frac{ \partial \Phi_{\bf w^\alpha} } { \partial \delta_i^{\alpha} (x_i) } 
= -\mu_\alpha (x_i)
&= - \sum_{ x_{\alpha \backslash i} } \mu_\alpha (x_\alpha) 
= - \sum_{x_c} \cdots \sum_{x_{i+1}} \prod_{j=i}^{c} \mu_\alpha (x_j | x_{j+1:c} )
\end{align}
In addition, the derivative of $\Phi_{\bf w^\alpha}$ w.r.t. $w_i^\alpha$ is the conditional entropy,
\begin{align}
\label{grdt_w_i_alpha}
\frac{ \partial \Phi_{\bf w^\alpha} } { \partial w_i^{\alpha} } = H(x_i | x_{i+1:c}; \mu_\alpha(x_\alpha) )
= - \sum_{x_\alpha}  \mu_\alpha(x_\alpha) \log \mu_\alpha(x_i | x_{i+1:c})
\end{align}

\begin{proof}
Denote the reparameterization on clique $\alpha$ as
$
\widetilde{\theta}_{\alpha} (x_\alpha) = \theta_{\alpha} (x_\alpha) -\sum_{i\in \alpha} \delta_i^{\alpha}(x_i).
$\\
From the recursive definition of $Z_i (x_{i+1:c})$ \eqref{recur_Zi}, we have the following recursive rule for gradient,
\begin{align}
\frac{\partial \log Z_i (x_{i+1 : c}) } {\partial \widetilde{\theta}_{\alpha} (x_\alpha)} 
&= \frac{\partial } {\partial \widetilde{\theta}_{\alpha} (x_\alpha)} 
\Big(  w_i^\alpha \log \sum_{x_i} \big[ Z_{i-1}(x_{i:c}) \big]^{1/w_i^\alpha} \Big) 	\nonumber
\\&= w_i^\alpha \cdot \frac{  \frac{1}{w_i^\alpha}   \cdot   Z_{i-1}(x_{i:c})^{ \frac{1}{w_i^\alpha}} } 
												{ \sum_{x_i} \big[ Z_{i-1} (x_{i:c}) \big]^{\frac{1}{w_c^\alpha}}  } 
 \cdot   Z_{i-1}(x_{i:c})^{ -1}   \cdot   \frac{\partial Z_{i-1}(x_{i:c}) } { \partial \widetilde{\theta}_{\alpha} (x_\alpha)  } \nonumber
%
%
\\&= \mu_\alpha (x_i | x_{i+1:c}) \cdot \frac{\partial \log Z_{i-1} (x_{i:c})  } { \partial \widetilde{\theta}_{\alpha} (x_\alpha) }.
\label{recur_derivative_logZi}
\end{align}
It should be noted, implicitly,  $x_{i+1 : c}$ within $ \widetilde{\theta}_{\alpha} (x_\alpha)$ should take the same value as
$x_{i+1 : c}$ in $\log Z_i (x_{i+1 : c})$, otherwise, the derivative will equal $0$.

As a result, we can calculate the derivatives of $\Phi_{\bf w^\alpha} (\widetilde{\theta}_{\alpha} )$
 w.r.t. $\widetilde{\theta}_{\alpha} (x_\alpha)$ recursively as,
\begin{align}
\label{grdt_theta_bar}
\frac{ \partial \Phi_{\bf w^\alpha} } { \partial \widetilde{\theta}_{\alpha} (x_\alpha) } 
& = \frac{ \partial \log Z_c } { \partial \widetilde{\theta}_{\alpha} (x_\alpha) }
= \mu_\alpha(x_c) \cdot \frac{\partial \log Z_{c-1} (x_c) } {\partial \widetilde{\theta}_{\alpha} (x_\alpha)  }
= \cdots
= \prod_{i=1}^c \mu_\alpha (x_i | x_{i+1 : c}) =  \mu_\alpha (x_\alpha).
\end{align}
By the chain rule,
\begin{align*}
\frac{ \partial \Phi_{\bf w^\alpha} } { \partial \delta_i^{\alpha} (x_i)  } 
=  \sum_{ x_{\alpha \backslash i} } 
\frac{ \partial \Phi_{\bf w^\alpha} } { \partial \widetilde{\theta}_{\alpha} ( x_i,  x_{\alpha \backslash i} ) } \cdot
\frac{\partial \widetilde{\theta}_{\alpha} ( x_i,  x_{\alpha \backslash i} ) }  { \partial \delta_i^{\alpha} (x_i) } 
= - \sum_{ x_{\alpha \backslash i} } \mu_\alpha (x_\alpha),
\end{align*}
then \eqref{grdt_delta_i_alpha} has been proved.

Applying the variational form of powered-sum \eqref{exact_variational_form_wsum} to $\Phi_{\bf w^\alpha}$, we have 
$$
\Phi_{\bf w^\alpha}  ( \widetilde{\theta}_{\alpha} ) 
=  \max_{ b_\alpha \in \mathbb{M_\alpha} (G) }   \big \{ \langle \widetilde{\theta}_{\alpha} , b_\alpha \rangle 
		+ \sum_{i} w_i^\alpha H(x_i | x_{i+1 : n};  b_\alpha) \big \}.
$$
According to Danskin's theorem, the derivative
$
\frac{ \partial \Phi_{\bf w^\alpha} } { \partial \widetilde{\theta}_{\alpha} (x_\alpha) } 
= b_\alpha^* (x_\alpha) , 
$
which is the optimum of RHS. 
Combined with \eqref{grdt_theta_bar}, we have $b_\alpha^* = \mu_\alpha$ immediately, and 
 the derivative  w.r.t. $ w_i^{\alpha}$ is,
\begin{align*}
\frac{ \partial \Phi_{\bf w^\alpha} } { \partial w_i^{\alpha} } 
&= H(x_i | x_{i+1:c}; \mu_\alpha (x_\alpha)),
\end{align*}
then \eqref{grdt_w_i_alpha} has been proved.

\end{proof}
\end{pro}

\end{document}